%% file: main.tex
\documentclass[10pt,twocolumn,letterpaper]{article}

\usepackage[pagenumbers]{cvpr} 

\input{preamble}

\usepackage{algorithm}
\usepackage{algpseudocode}
\definecolor{cvprblue}{rgb}{0.21,0.49,0.74}
\usepackage[pagebackref,breaklinks,colorlinks,citecolor=cvprblue]{hyperref}

\def\paperID{7361} 
\def\confName{CVPR}
\def\confYear{2024}

\title{SCoFT: Self-Contrastive Fine-Tuning for Equitable Image Generation}
\author{Zhixuan Liu$^{1}$ \and Peter Schaldenbrand$^{1}$ \and Beverley-Claire Okogwu$^{1}$ \and Wenxuan Peng$^{2}$ \and Youngsik Yun$^{3}$ \quad Andrew Hundt$^{1}$ \quad Jihie Kim$^{3}$ \quad Jean Oh$^{1}$ \\
$^1$Carnegie Mellon University \quad
    $^2$Nanyang Technological University \quad  
    $^3$ Dongguk University\\
}

\begin{document}
\twocolumn[{%
\renewcommand\twocolumn[1][]{#1}%
\maketitle
\begin{center}
    \vspace{-0.5cm}
    \centering \small
    \captionsetup{type=figure}
    \includegraphics[width=0.91\textwidth]{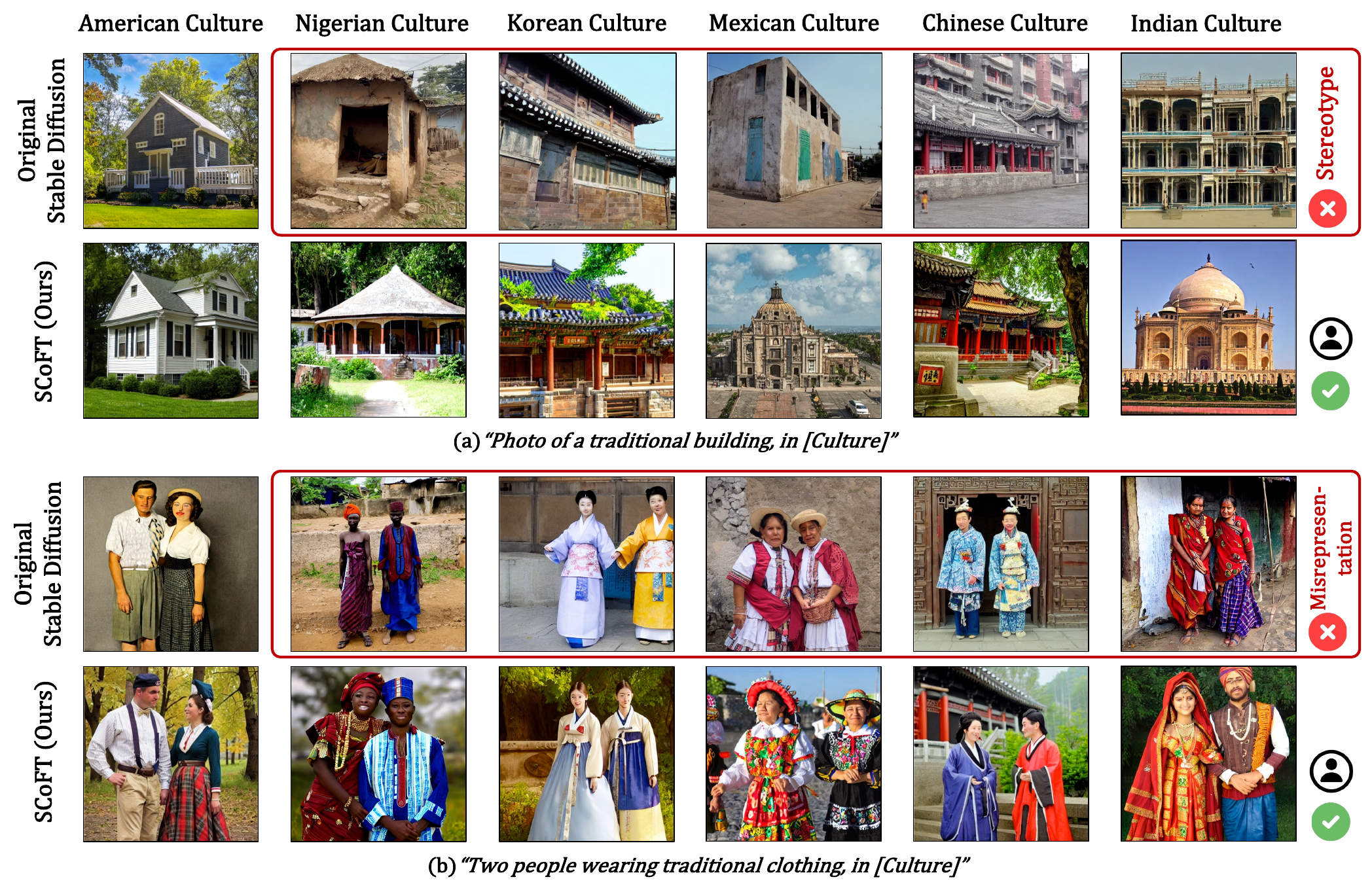}\vspace{-13pt}%
    \captionof{figure}{Comparison between Stable Diffusion with and without our proposed fine-tuning approach, SCoFT, on our proposed CCUB dataset.  Stable Diffusion perpetuates harmful stereotypes that assume dirty buildings are representative of some nations, and often generates regionally irrelevant designs. By contrast, our approach decreases stereotypes and improves cultural relevance of generated images.}%
    \label{fig:teaser}%
\end{center}%
}]
\input{sec/0_abstract}    
\input{sec/1_intro}

\input{sec/2_related_work}
\input{sec/4_dataset}%
\input{sec/5_method_new}
\input{sec/6_experiments_new}

\input{sec/7_results}

\input{sec/8_conclusion}
\input{sec/9_acknowledgements}
{
    \small
    \bibliographystyle{ieeenat_fullname}
    \bibliography{main}
}
\input{sec/X_suppl}

\end{document}

%% file: preamble.tex
\usepackage[dvipsnames]{xcolor}


%% file: sec/0_abstract.tex
\begin{abstract}
\vspace{-0.3cm}
%
Accurate representation in media is known to improve the well-being of the people who consume it. 
Generative image models trained on large web-crawled datasets such as LAION are known to produce images with harmful stereotypes and misrepresentations of cultures. 
We improve inclusive representation in generated images by (1) engaging with communities to collect a culturally representative dataset that we call the Cross-Cultural Understanding Benchmark (CCUB) and (2) proposing a novel Self-Contrastive Fine-Tuning (SCoFT) method that leverages the model's known biases to self-improve. 
SCoFT is designed to prevent overfitting on small datasets, encode only high-level information from the data, and shift the generated distribution away from misrepresentations encoded in a pretrained model.
Our user study conducted on 51 participants from 5 different countries based on their self-selected national cultural affiliation shows that fine-tuning on CCUB consistently generates images with higher cultural relevance and fewer stereotypes when compared to the Stable Diffusion baseline, which is further improved with our SCoFT technique. Resources can be found at \url{https://ariannaliu.github.io/SCoFT}.%
\end{abstract}%

%% file: sec/1_intro.tex

\vspace{-0.7cm}%
\section{Introduction}
\vspace{-0.2cm}
\label{sec:intro}
Representation matters.
In media, studies repeatedly show that representation affects the well-being of its viewers~\cite{shaw2010representationInVideoGames,caswell2017representation,elbaba2019teenMediaRepr}.  
Representation can positively affect viewers by providing them with role models that they identify with, but it can also negatively affect viewers by creating harmful, stereotypical understandings of people and culture~\cite{castaneda2018powerRepresentation}.
When people are accurately represented in media, it allows people to properly understand cultures without harmful stereotypes forming~\cite{dixon2000overrepresentation,mastro2000portrayal}.
To date, unfortunately, many media-generating AI models show poor representation in their results~\cite{ntoutsi2020bias,luccioni2023stable} and have been deployed for cases with negative impacts on various groups, such as nonconsensual images~\cite{fourzeromedia2023marketplace}.  
Many of these issues stem from their large training datasets which are gathered by crawling the Internet without filtering supervision and contain malignant stereotypes and ethnic slurs among other problematic content~\cite{birhane2021stereotypesInLAION}, impacting a range of applications\cite{hundt2022robots_enact}. 
Researchers have shown that large datasets such as LAION-400M~\cite{schuhmann2021laion} used to train many text-to-image synthesis models, including Stable Diffusion~\cite{rombach2021stableDiffusionOriginal}, center the Global North~\cite{birhane2021stereotypesInLAION,birhane2023hate,luccioni2023stable} and struggle to accurately depict cultures from the Global South as shown in Figure \ref{fig:teaser}.

To ensure models better represent culture and more accurately represent the world, we introduce a new task of \textit{culturally-aware} image synthesis with the aim of addressing representation in image generation: generating visual content that is perceived to be more representative of national cultural contexts. Our overarching goal is to improve the well-being of viewers of the AI-generated images with particular attention to those who are from a selection of groups marginalized by existing methods. 
Our research question is, how can effective, existing text-to-image models be improved to become more culturally representative and thus less offensive? Since it may be infeasible to vet billions of training examples for accurate cultural content,
we hypothesize that a small dataset that is veritably representative of a culture can be used to prime pre-trained text-to-image models to guide the model toward more culturally accurate content creation.  To verify the hypothesis, we collected a dataset of image and caption pairs for 5 cultures.  For each culture, data was collected by people who self-selectedly affiliated with that culture as they are the people who properly understand it and are most affected by its misrepresentations. We call this the Cross-Cultural Understanding Benchmark (CCUB) dataset which comprises 150 - 200 images for each culture each with a manually written caption as shown in Figure~\ref{fig:dataset}.


To encode the culturally representative information in CCUB into a pre-trained model, we propose to fine-tune the model with the new dataset.
Existing fine-tuning techniques work well for low-level adaptations such as style changes or introducing new characters to models~\cite{hu2021lora}, but we show that these methods struggle to encode high-level, complex concepts such as culture. Additionally, fine-tuning on small datasets, such as CCUB, can lead to overfitting.

Unlike concept editing tasks~\cite{Gandikota_2024_WACV, gandikota2023concept} with specific image editing directions, depicting cultural accuracy remains more abstract and challenging.
We propose a novel fine-tuning approach, Self-Contrastive Fine-Tuning (SCoFT, pronounced /sôft/), to address these issues. SCoFT leverages the pre-trained model's cultural misrepresentations against itself. We harness the intrinsic biases of large pre-trained models as a rich source of counterexamples; shifting away from these biases gives the model direction towards more accurate cultural concepts.
Image samples from the pre-trained model are used as negative examples, and CCUB images are used as positive examples, to train the model to discern subtle differences. 
We de-noise latent codes in several iterations, project them into the pixel space, and then compute the contrastive loss. The loss is backpropagated to the diffusion model UNet and optimized to push generated samples towards the positive distribution and away from the negative.
This is all done in a perceptual feature space so that the model learns more high-level, complex features from the images.

To evaluate our results we recruited participants who identify as being a member of the cultural communities in the CCUB dataset to rank images generated by Stable Diffusion with or without the proposed self-contrastive fine-tuning on CCUB. Fine-tuning on CCUB was found to decrease offensiveness and increase the cultural relevance of generated results based on 51 participants across five cultures and 766 image comparisons. Our proposed SCoFT approach further improved these results. 
We share the findings from our experiments to provide a basis for an important aspect of AI-generated imagery: that cultural information should be accurately presented and celebrated equitably. Our contributions are as follows:
\begin{enumerate}
    \item The introduction of culturally-aware text-to-image synthesis as a valuable task within text-to-image synthesis;
    \item The Cross-Cultural Understanding Benchmark (CCUB) dataset of culturally representative image-text pairs across 5 countries; and 
    \item Self-Contrastive Fine-Tuning (SCoFT), a novel technique for encoding high-level information into a pre-trained model using small datasets.
\end{enumerate}

%% file: sec/2_related_work.tex
\section{Related Work}
%
\noindent\textbf{Cultural Datasets}
Various efforts have been made to evaluate and document the impacts of datasets\cite{scheuerman2021datasetpolitics,hutchinson2021mldatasetaccountability,wang2022measurecaptionharms}. 
Dollar Street~\cite{Rojas2022TheDS} aimed to capture accurate demographic information based on socioeconomic features, such as everyday household items and monthly income, of 63 countries worldwide. 
However, this dataset offers less diverse scenarios, as most of its images are indoor views with limited cultural features.
Likewise, the WIT~\cite{srinivasan2021wit} and MLM~\cite{armitage2020mlm} strive for cultural representation but use Wikipedia/WikiData sources for images that are not representative of all aspects of culture and are over-saturated with old, streotypical images.
Other works, capturing the idea of a diverse dataset, aim to reduce stereotypical bias through self-curation~\cite{desai2021redcaps} or cultural-driven methods~\cite{mandal2021dataset, miquel2019wikipedia, garcia2023uncurated}, inspiring our data collection methodology.

The MaRVL dataset~\cite{Liu2021VisuallyGR}, for example, was curated by people who identify as affiliated with one of several particular cultures,
MaRVL was developed to mitigate biases for reasoning tasks covering popular concepts and is unsuitable for text-to-image synthesis. 
Our dataset was also designed to cover a diverse sample of cultures and engage with people who are native, but it is specifically curated for vision-language tasks and diverse cultural categories.

\noindent\textbf{Fine-Tuning Text-to-Image Models}
%
%
Fine-tuning pre-trained text-to-image synthesis models with a new dataset is an approach to encode additional new simple concepts and content into the model~\cite{ruiz2023dreambooth, hu2021lora, TextToPokemonGenerator, japanse2022stableDiffusion, lu2023specialistDiffusion}. But culture is a complex, high-level concept that poses many challenges when attempting to fine-tune a model to understand it.

\noindent\textbf{Fine-Tuning on Small Datasets.}
Our CCUB dataset and, more generally, datasets that are collected and verified by hand rather than scraped from the internet en masse are small, leading to challenges when fine-tuning~\cite{moon2022fineTuneSmall}.  While fine-tuning techniques, such as LoRA~\cite{hu2021lora}, DreamBooth~\cite{ruiz2023dreambooth}, or Specialist Diffusion~\cite{lu2023specialistDiffusion}, adapt models to learn simple concepts (e.g., a single character's appearance) with only a hand full of training images, these techniques may overfit to the training data when fine-tuning for a high-level, complex concept such as a culture.

\noindent\textbf{Fine-Tuning on Cultural Data.}
Prior work in adapting pre-trained models to be culturally relevant found some success using millions of culturally relevant text-image pairs, as in ERNIE-ViLG 2.0.~\cite{feng2022ernie} and Japanese Stable Diffusion~\cite{japanse2022stableDiffusion}.  The size of these training datasets leads to better cultural representations of Japan and China, but it is not easy to be used universally, as these approaches require millions of training examples which cannot possibly be met for cultures with less internet presence. Besides, these datasets are so large that it is infeasible to vet them for harmful and stereotypical information. We propose a fine-tuning technique that adapts pre-trained models to learn complex, elusive concepts, namely culture, from small datasets. 

\noindent\textbf{Fine-Tuning Stable Diffusion in the Pixel Space.}
Latent diffusion models are customarily trained in the latent space, however, the latent codes can be decoded into images differentiably. Multiple works compute losses on the decoded images to optimize over the input latent code~\cite{ wallace2023doodle} or the decoder weights~\cite{avrahami2023blended}.
DiffusionCLIP~\cite{kim2022diffusionclip} optimizes UNet parameters using losses computed on image outputs, however, this is performed on a small, non-latent diffusion model. Latent diffusion models' high parameter count makes it intractable to record gradients through multiple passes of the UNet and then decode into the pixel space. We develop a novel method for reducing the size of the computation graph to make backpropagating loss through Stable Diffusion tractable.

\noindent\textbf{Perceptual Losses.}
Perceptual losses, such as LPIPS~\cite{zhang2018LPIPS}, have been shown to align more closely with human perception than pixel space Euclidean distance~\cite{schaldenbrand2023frida}. In this work, we use perceptual loss to prevent over-fitting on our small CCUB data, because perceptual losses ignore low-level differences (e.g., colors) and capture high-level details (e.g., objects), which are more important for the complex concepts such as cultures. To our knowledge, no other work has trained a latent diffusion model using perceptual loss, likely due to the technical challenges of back-propagating loss through the diffusion process and latent decoder, which we address in this paper.

\noindent\textbf{Contrastive Losses.}
Perceptual features have also been used in contrastive losses~\cite{manocha2021audioContrast}, 
where a model is trained to generate samples that are similar in distribution to a dataset. We build on these approaches and introduce a Self-Contrastive method that uses data produced from a pre-trained model as negative examples along with a veritable dataset as positive examples to push the model towards producing from the positive distribution. Utilizing the model's inherent counterexamples facilitates the identification of abstract and complex concepts, such as culture, which are typically challenging to articulate.

%% file: sec/4_dataset.tex
\begin{figure}[t]
  \centering
   \includegraphics[width=0.9\linewidth]{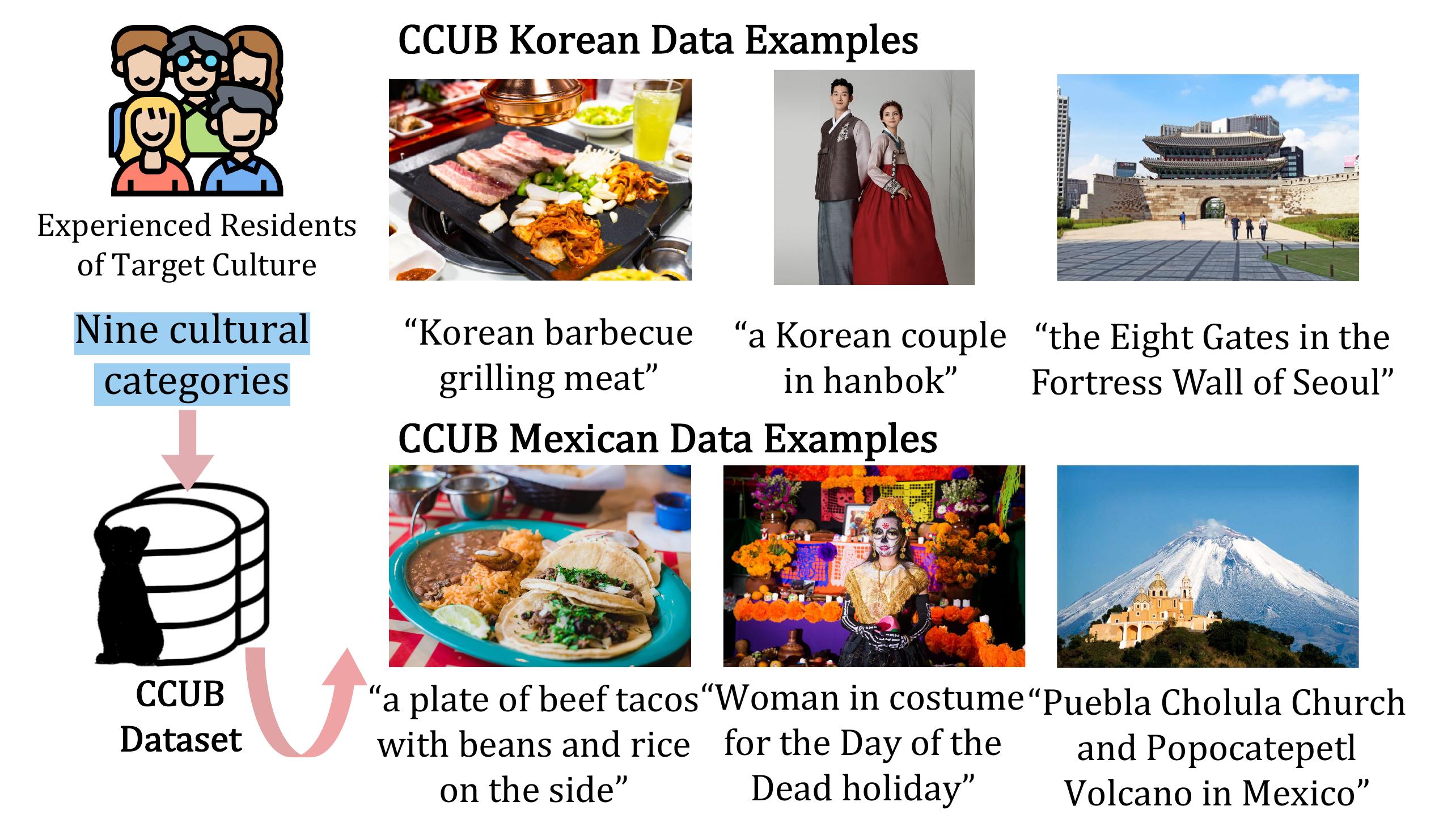}\vspace{-10pt}%
   \caption{Sample cultural images and captions from our proposed CCUB dataset.}%
   \label{fig:dataset}\vspace{-15pt}%
\end{figure}%

\section{CCUB Dataset}
\textit{Can a CCUB tame a LAION?} As opposed to the LAION~\cite{schuhmann2022laion} dataset which scraped images and captions from the internet with minimal supervision leading to a prominence in harmful content, our CCUB dataset was collected by hand by the people most affected by cultural misrepresentations in text-to-image synthesis.

Following the definition of culture in~\cite{10.2307/2378980},~\cite{Liu2021VisuallyGR}, and~\cite{ids}, nine categories are used to represent cultural elements in our dataset: architecture (interior and exterior), city, clothing, dance music and visual arts, 
food and drink, nature, people and action, religion and festival, utensils and tools. 
The categories are further divided into traditional and modern to reflect how cultural characteristics change over time.

For each culture, we recruited at least 5 people with at least 5 years of experience living in one of 5 countries (China, Korea, India, Mexico, and Nigeria) to each provide 20-30 images and captions. 
The images were collected either by collecting image links from Google searches or the collectors' own photographs.
%
Figure \ref{fig:dataset} shows some selected samples of our CCUB dataset.

%% file: sec/5_method_new.tex
\begin{figure*}[t]
  \centering \small
   \includegraphics[width=0.95\linewidth]{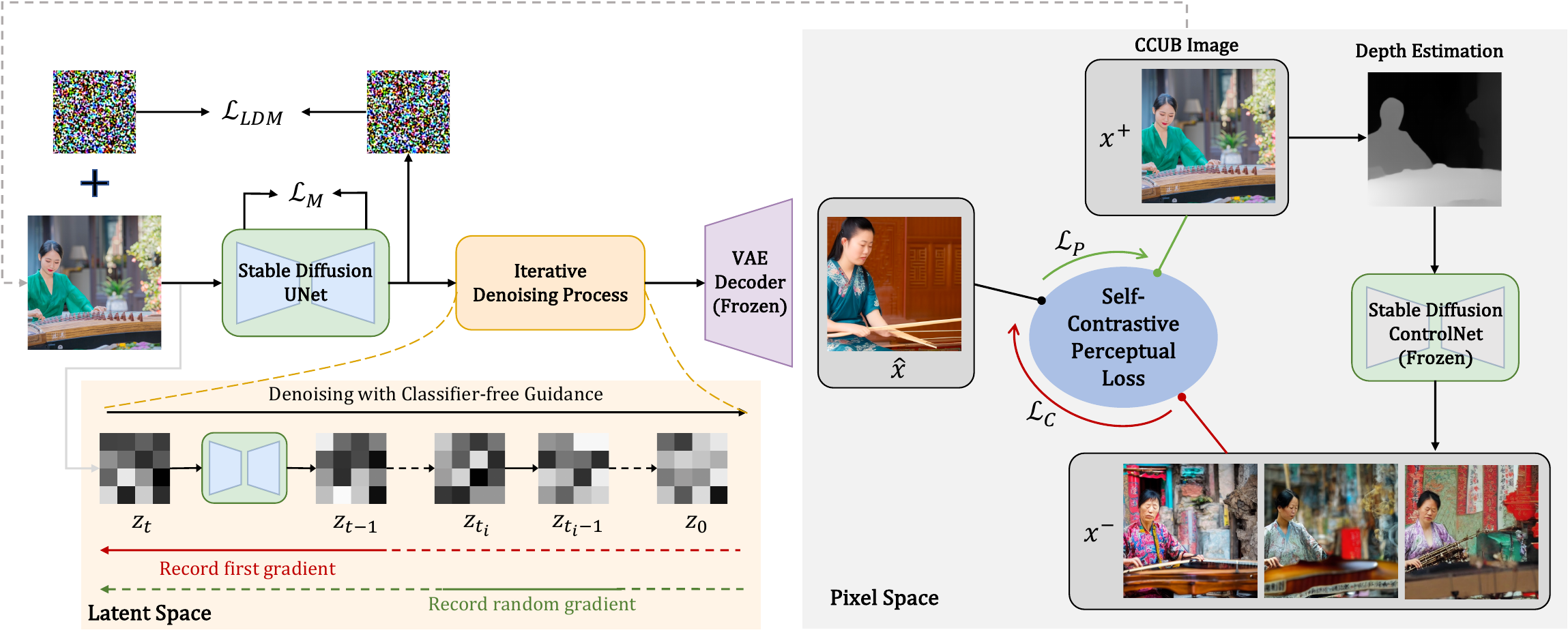}\vspace{-10pt}
   \caption{\textbf{SCoFT Overview.} A conventional fine-tuning loss, $\mathcal{L}_{LDM}$, and memorization penalty loss, $\mathcal{L}_{M}$, are computed in the Stable Diffusion latent space using images and captions from our CCUB dataset.  After 20 denoising steps, the latent space is decoded. Perceptual features are extracted from the generated image and compared contrastively to CCUB images as positive and non-fined-tuned Stable Diffusion images as negative examples to form our Self-Contrastive Perceptual Loss, ${\mathcal{L}_{C}}$. }%
   \label{fig:overview}%
   \vspace{-0.5cm}%
\end{figure*}
\vspace{-0.2cm}
\section{Method}
\vspace{-0.2cm}
Given our CCUB dataset, our objective is to alter Stable Diffusion to have a more accurate understanding of a given culture.
We next offer a brief background on training latent diffusion models (Sec.~\ref{subsec:background}), a modification to a regularization loss to prevent over-fitting (Sec.~\ref{subsec:memorization_loss}), a novel approach to computing perceptual loss on decoded images (Sec.~\ref{subsec:perceptual_loss}), and a method to contrastively use Stable Diffusion's misrepresentations of culture to refine itself (Sec.~\ref{subsec:contrastive_loss}).
%

\subsection{Latent Diffusion Model Loss}
\label{subsec:background}
Diffusion models are latent variable models that sample from a desired distribution by reversing a forward noising process.
The latent noise at any timestep $t$ is given by $\mathbf{z}_t = \sqrt{\alpha_t}\mathbf{z}_0 + \sqrt{1-\alpha_t}\epsilon$, where $\mathbf{z}_0$ is the latent variable encoded from the real image and $\alpha_t$ is the strength of the gaussian noise $\epsilon$. We are interested in the pretrained deonising network $\epsilon_\theta(\mathbf{z}_t,\mathbf{c},t)$, which denoises the noisy latent to get $\mathbf{z}_{t-1}$ conditioned on the text $\mathbf{c}$. The training objective for the denoiser $\epsilon_\theta$ is minimizing the noise prediction:
\begin{equation}\label{eq:1}
\mathcal{L}_{LDM}(\mathbf{z}, \mathbf{c})=\mathbb{E}_{\epsilon, \mathbf{z}, \mathbf{c}, t}\left[w_t\left\|\epsilon-\epsilon_\theta\left(\mathbf{z}_t, \mathbf{c}, t\right)\right\|\right]
\end{equation}
where $w_t$ is the weight to control the noise schedules. 
%
A pretrained model can be further trained, or fine-tuned, using the original objective with a new dataset.  In this paper, we use the Low-Rank Adaptation (LoRA)~\cite{hu2021lora} approach which reduces memory demands and over-fitting by training new low-rank weight matrices.

\subsection{Memorization Loss}\label{subsec:memorization_loss}
Despite CCUB's rich cultural content, its size still remains small compared to LAION. The challenges of language drift and decreased output diversity are prevalent issues encountered during this few-shot fine-tuning \cite{ruiz2022dreambooth}. Moreover, CCUB's text captions can be highly specific, as shown in Fig.~\ref{fig:dataset}. Fine-tuning solely on the CCUB dataset using Equation \ref{eq:1} may lead to the undesirable outcome of reproducing training data as shown in Figure~\ref{fig:overfitting}. Inspired by \cite{kumari2023conceptablation}, which proposed a model-based concept ablation by letting the model memorize the mapping between newly generated anchor images $\textbf{x}^{\varnothing}$ and $\mathbf{c}^{*}$, our approach differs by focusing on preventing memorizing during fine-tuning on a small dataset (e.g., CCUB). We harness  the property of BLIP automatic caption $\mathbf{c}_{blip}$ on $\mathbf{x}_{ccub}$, which is the naive version of our cultural text prompt $\mathbf{c}_{ccub}$, to regularize the model outputs conditioned on $\mathbf{c}_{ccub}$, as shown in Figure \ref{fig:overfitting}.
To achieve this, we introduce a memorization penalty loss leveraging BLIP~\cite{li2022blip} generated captions of CCUB images. We utilize multiple BLIP captions $\{\mathbf{c}_{blip}\}$ to regularize the one-on-one mapping between CCUB images and cultural captions:
\begin{align}\label{eq:diversity}
\begin{split}
\mathcal{L}_{M} (\mathbf{x}_{ccub}, \mathbf{c}_{ccub}, \mathbf{c}_{blip}) = \mathbb{E}_{\epsilon,t}[ \|{\epsilon}(\mathbf{x}_{ccub}, \mathbf{c}_{ccub}, t) \\
 - \mathbb{E}_i[{\epsilon}(\mathbf{x}_{ccub}, \mathbf{c}^i_{blip}, t). \operatorname{sg}()]\|],
\end{split}
\end{align}
where $. \operatorname{sg}()$ stands for a stop-gradient operation of the current network to reduce memory cost.

\subsection{Perceptual Loss}\label{subsec:perceptual_loss}
$\mathcal{L}_{\text{LDM}} $ and $\mathcal{L}_{M}$ are loss functions that operate on the latent codes within a diffusion model. Operating directly on these latent codes is ideal for fine-tuning models for adding simple concepts such as adding a character's appearance to the model. For adding more complex, abstract concepts, like culture, we propose to decode the latent space into the pixel space in order to utilize pre-trained perceptual similarity models.  We propose to use a perceptual loss, $\mathcal{L}_{P}$, which is computed as the difference in extracted perceptual features between the decoded, generated image, $\hat{x}$, and an image from a training set, $x$:
\begin{align}
\mathcal{L}_{P}(\hat{x}, x)
= \mathbb{E}_{\hat{x}, x}[\mathcal{S}(\hat{x}, {x};f_{\theta}) ] \label{eq:perceptual_loss}
\end{align}
where $\mathcal{S}$ is some perceptual similarity function and $f_{\theta}$ is a pretrained feature extractor.

\noindent\textbf{Backpropagation through sampling.}
State-of-the-art perceptual models typically process inputs in the pixel space. 
In contrast, Stable Diffusion is fine-tuned in a latent space.
To fulfill our objective function, an intuitive strategy entails iteratively denoising latent features and then decoding them back into the pixel space for use with perceptual models. Concurrent work \cite{clark2023directly} denoises the stable diffusion latent from Gaussian noise into the pixel space based solely on text prompts and by optimizing the Stable Diffusion UNet with a reward function computed using the decoded image. 
Instead, our approach starts from the latent code at timestep $t$: $\mathbf{z}_t = \sqrt{\alpha_t}\mathbf{z}_0 + \sqrt{1-\alpha_t}\epsilon$, making it coupled with the Stable Diffusion fine-tuning process. 
We utilize classifier-free guidance, iteratively denoising the latent code according to $(1+w(t)) \epsilon\left(\mathbf{z}_t, \mathbf{c}_{ccub}, t\right)-w(t) \epsilon\left(\mathbf{x}_t, \varnothing, t\right)$, where $w(t)$ is the guidance scale and $\varnothing$ is a null text embedding. 
This enables the model to generate images conditioned on cultural text prompts and unveils its cultural understanding, as illustrated in Figure \ref{fig:overview}. In practice, we denoise $\mathbf{z}_t$ for 20 timesteps.
Starting from $\mathbf{z}_t$, our method ensures that the denoised image aligns with the same pose and structure as the original training image $\textbf{x}_0$. This facilitates a more meaningful comparison for perceptual loss and subsequent self-contrastive perceptual loss, exposing major differences in cultural concepts.

Directly backpropagating through the multiple UNet denoising iterations, the latent space decoder, and the perceptual model produce huge memory costs and computational time. To address this, we selectively record the gradient on a single denoising step and employ stopgrad on other denoising steps. Our findings indicate that recording the gradient from the first step has the most significant impact on refining the model's cultural understanding. Further comparisons on gradient recording and perceptual model backbone are detailed in Sec.~\ref{sec:perceptual_and_gradient}.


\subsection{Self-Contrastive Perceptual Loss} \label{subsec:contrastive_loss}
To further improve Perceptual Loss, we raise an intriguing question: Can Stable Diffusion leverage its intrinsic biases to refine its own? We seek to leverage the model's prior of its cultural understanding and propose a contrastive learning approach.
Utilizing our CCUB dataset, we designate the positive examples to be $\{\mathbf{x}^+ | \mathbf{x}_{ccub}\}$ representing preferred features. 
To unveil the cultural biases within Stable Diffusion, we employ images generated by the model itself as negative samples.

It is imperative to ensure that the generated negative samples share a high-level
similarity with positive samples, such as pose and structure, thereby emphasizing that the primary distinctions lie in the diffusion model's perception of cultural features. 
We achieve this by incorporating a pre-trained ControlNet~\cite{zhang2023controlnet} module $\Theta_c$, conditioned on the estimated depth of $\{\mathbf{x}^+\}$, into Stable Diffusion. As depicted in Figure~\ref{fig:overview}, negative examples are obtained as  $\{\mathbf{x}^-_i | \Theta_c(\mathcal{D}(\mathbf{x}^+), \mathbf{c}) \}$, where $\mathcal{D}$ is the MiDaS~\cite{birkl2023midas, Ranftl2022midasPaper} depth estimator, and $\mathbf{c}$ represents $\mathbf{c}_{blip}$ followed by a cultural suffix (e.g., ``in Korea"). 

To enhance the cultural fine-tuning process, our objective is to ensure that images generated by the current model, denoted as $\hat{x}^0_{\theta_t}$ have closer perceptual distances to positive examples and farther distances from negative examples. This reinforces the model's alignment with preferred cultural features, distancing itself from undesirable biases in negative examples, indicated by: $\mathcal{S}(\hat{x}^0_{\theta_t}, {x}^+;f_{\theta}) >  \mathcal{S}(\hat{x}^0_{\theta_t}, {x}_i^-;f_{\theta})$, where $\mathcal{S}$ is some perceptual similarity function and $f_{\theta}$ is a pre-trained feature extractor.  Thus, we formulate this objective, which we call  Self-Contrastive Perceptual Loss, using triplet loss:
\begin{equation}\label{eq:contrastive}
\begin{split}
\mathcal{L}_{C}(\hat{x}, x^+, x^-)
= \mathbb{E}_{\hat{x}, x^+, x^-}[\max(\mathcal{S}(\hat{x}, {x}^+;f_{\theta}) \\
-\lambda \mathcal{S}(\hat{x}, {x}^-;f_{\theta}) + m, 0)]
\end{split}
\end{equation}
where $\lambda$ denotes the weights on negative examples, $m$ is the constant margin between positive and negative examples, and $f_\theta$ is a feature extractor. We evaluate a variety of state-of-the-art perceptual embeddings and report comparison results in Section~\ref{sec:perceptual_and_gradient}.

Our full fine-tuning approach, SCoFT, is a weighted sum of all previously mentioned loss functions.
~\\

%% file: sec/6_experiments_new.tex
\vspace{-20pt}%
\section{Experiments}%
\begin{figure*}[t]
  \centering
   \includegraphics[width=1\linewidth]{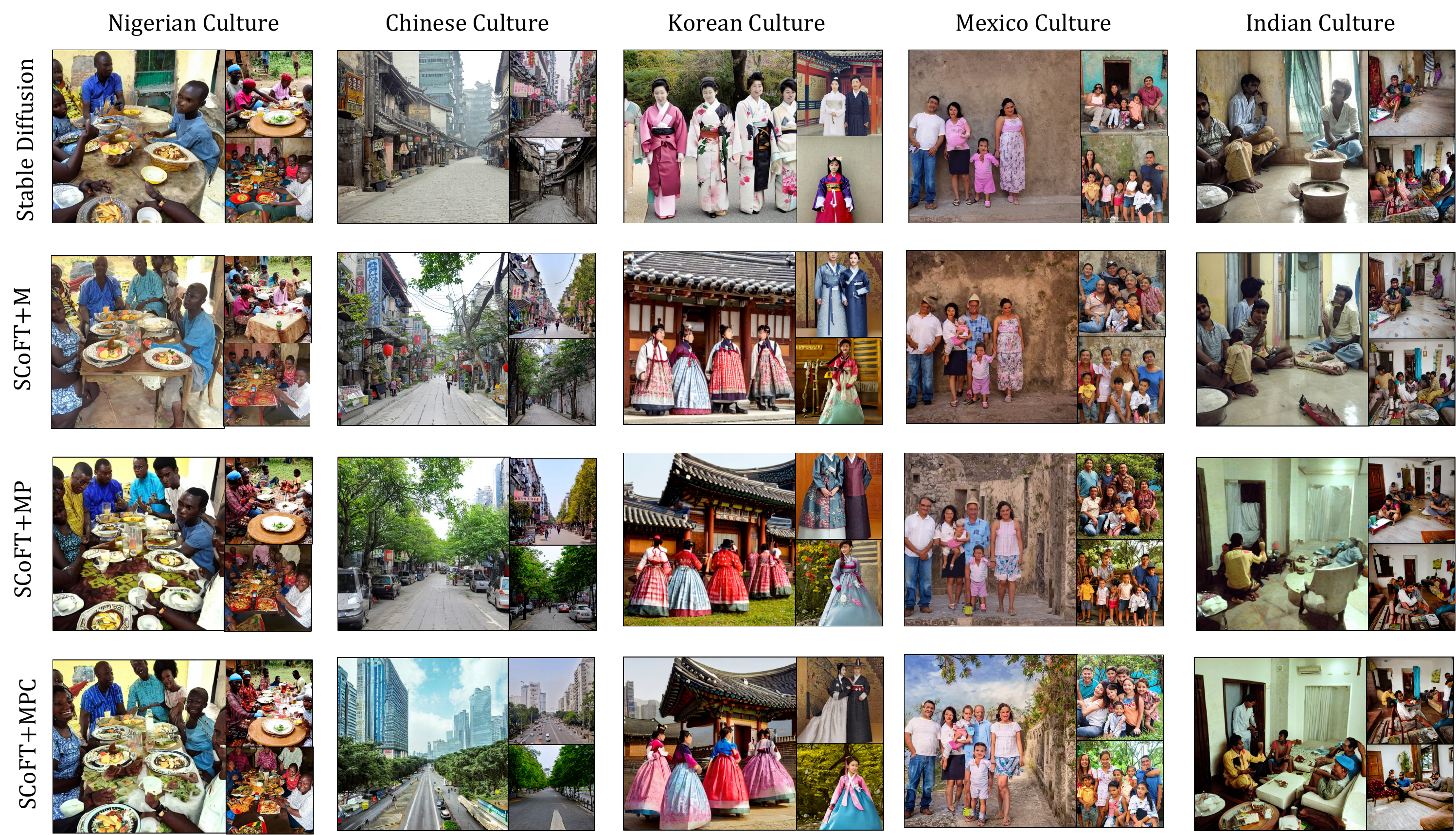}\vspace{-10pt}%
   \caption{Qualitative comparison of our SCoFT model ablated and compared to Stable Diffusion without fine-tuning.}%
   \label{fig:qualitative_ablation}%
   \vspace{-0.1cm}%
\end{figure*}

\begin{table*}[h!]
    \small
    \centering
    \begin{tabular}{l|p{.2cm}p{.2cm}p{.3cm}||ccc||cccc}
    \toprule
     \multicolumn{4}{c||}{} &
     \multicolumn{3}{c||}{Automatic Metrics} &
     \multicolumn{3}{c}{User Survey Results - Average Rank} 
     \\
    \hline
    Model Name 
    & $\mathcal{L}_{{M}}$ 
    & $\mathcal{L}_{P}$ 
    & $\mathcal{L}_{C}$ 
    & \begin{tabular}[c]{@{}c@{}}KID- $\downarrow$\\CCUB\end{tabular}
    & \begin{tabular}[c]{@{}c@{}}KID- $\downarrow$\\COCO\end{tabular} 
    & \begin{tabular}[c]{@{}c@{}}CLIP- $\uparrow$\\ Score\end{tabular} 
    &\begin{tabular}[c]{@{}c@{}}Best \\ Described\end{tabular} 
    & \begin{tabular}[c]{@{}c@{}}Most Culturally\\ Representative\end{tabular}
    &  \begin{tabular}[c]{@{}c@{}}Least \\ Stereotypical\end{tabular}  
    &  \begin{tabular}[c]{@{}c@{}}Least \\ Offensive\end{tabular}  \\ \midrule
    Stable Diffusion\cite{rombach2021stableDiffusionOriginal} & & & 
    & 30.355 & \textbf{4.396} & \textbf{0.813}
    &  3.09 &  3.07 & 2.98 &  3.07        \\
    SCoFT+M     & \checkmark  &    &   
    & 22.643 & 4.711 & 0.802
    &  2.57   & 2.56 & 2.66 & 2.59   \\
    SCoFT+MP    & \checkmark  & \checkmark    & 
    & 21.360 & 4.936 & 0.800 
    & 2.33  & 2.35 &2.34 &  2.30   \\
    SCoFT+MPC   & \checkmark  &  \checkmark  &  \checkmark  
    & \textbf{19.621} & 4.819 & 0.799 
    &  \textbf{1.83} & \textbf{1.84} & \textbf{1.91} & \textbf{1.78} \\
    \bottomrule
    \end{tabular}
    \caption{We compare our SCoFT ablations to Stable Diffusion using automatic metrics (Sec.~\ref{sec:automatic_results}) and a user survey (Sec.~\ref{sec:user_survey}). Values in the user survey results report average ranking of images across all five cultures where lower rankings indicate better results (KID is $\times 10^3$) }
    \label{tab:survey}
    \vspace{-0.25cm}
\end{table*}
\noindent\textbf{User Survey.}\label{sec:user_survey}
Our goal of improving the cultural perception of generated images is a subjective metric largely determined by members of a given identity group. 
To evaluate our performance on this criterion, we recruited people with at least 5 years of cultural experience in each of the 5 countries with survey questions specific to their self-selected national cultural affiliation. 
A single page of the survey form provides one description (prompt) and one image made by  four different generators using a common random seed, for four total images.
We compare four image generators: Stable Diffusion against three fine-tuned ablations. All fine-tunings were performed with the CCUB dataset using $\mathcal{L}_{LDM}$ along with one or more proposed loss functions, see Tab.~\ref{tab:app_automatic_metrics}. For example, SCoFT+M is Stable Diffusion fine-tuned on CCUB using the sum of $\mathcal{L}_{LDM}$ and $\mathcal{L}_{M}$ as a loss function.
Each survey page has a total of four survey items (rows that participants respond to) to rank images on (a) Description and Image Alignment, (b) Cultural Representation, (c) Stereotypes, and (d) Offensiveness.
Participants respond by numerically ranking the set of randomly ordered images from best image to worst image once for each item.
An image labeled rank 1 would signify both best aligned and least offensive when each case is ranked, while rank 4 would be least well aligned and most offensive.
Details about our survey can be found in Sec.~\ref{subsec:amt_appendix}.

We quantitatively estimate the subjective perceived performance of each method with Matrix Mean-Subsequence-Reduced (MMSR)~\cite{NEURIPS2020_mmsr} model in crowd-kit~\cite{ustalov2023crowdkit}, an established algorithm~\cite{majdi2023crowdcertain} for noisy label aggregation, followed by a weighted majority vote to aggregate labels across workers, and then a simple majority vote aggregating labels into rankings, thus MMSR+Vote.

\noindent\textbf{Automatic Metrics.}
In addition to the user survey, we use Kernel Inception Distance (KID)~\cite{binkowski2018kid} and CLIP Score~\cite{hessel2021clipscore} to evaluate the quality of generated images. 
For automatic evaluation to ablate SCoFT, we use 10 test prompts for each culture, generating 20 images for each prompt.

%% file: sec/7_results.tex
\section{Results}
%
%
%
%

\noindent\textbf{Qualitative Comparison.}
We qualitatively compare our SCoFT model versus the original Stable Diffusion in Figure~\ref{fig:teaser}. SCoFT is able to guide Stable Diffusion away from generating stereotypes and misrepresentations of culture. For example, many of the Stable Diffusion results for a ``Photo of a traditional building, in ..." depict disheveled structures, which promote a harmful stereotype that some cultures are poor or simple, whereas SCoFT promotes more accurate and less stereotypical buildings for each nation.
To investigate the effects of each loss function within SCoFT we also qualitatively compare each ablation in Figure~\ref{fig:qualitative_ablation}. We tend to see the SCoFT models modernize generated images, which decreases harmful stereotypes.
%

\noindent\textbf{User Survey Results.} %
%
51 survey participants from five countries ranked images across four ablations by responding to each of the four survey items in Sec.~\ref{sec:user_survey}. 
The average participant rankings can be found in Table~\ref{tab:survey}. 
We also ran the MMSR~\cite{NEURIPS2020_mmsr} noisy data labeling algorithm across all responses (see Sec. \cref{sec:user_survey}, Supplement), 
finding a participant consensus ranking of: (1) SCoFT+MPC (2) SCoFT+MP, (3) SCoFT+M, and finally (4) Generic Stable Diffusion.
MMSR found that particpants reached an identical consensus when separately ranking each ablation with respect each of the four survey items.
MMSR also found a participant consensus in which every country individually agreed with the ranking above, with the exception of India, which swapped (1) SCoFT+MP and (2) SCoFT+MPC.


We convert the rankings into binary comparisons by isolating two ablations and comparing their rankings.  This way, we can compare the effects of each of the loss functions of SCoFT. SCoFT+M was ranked less offensive than Stable Diffusion 63\% of the time, SCoFT+MP was less offensive than SCoFT+M 56\% of the time, and SCoFT+MPC was less offensive than SCoFT+MP 62\% of the time. We see that each loss function contributed significantly to decreasing the offensiveness of generated results, and this trend continued for the other three survey items. We note that  the initial addition of fine-tuning and the contrastive loss produced more dramatic improvements in SCoFT compared to adding perceptual loss.

\begin{figure}
\vspace{-0.3cm}
    \centering
    \includegraphics[width=\columnwidth]{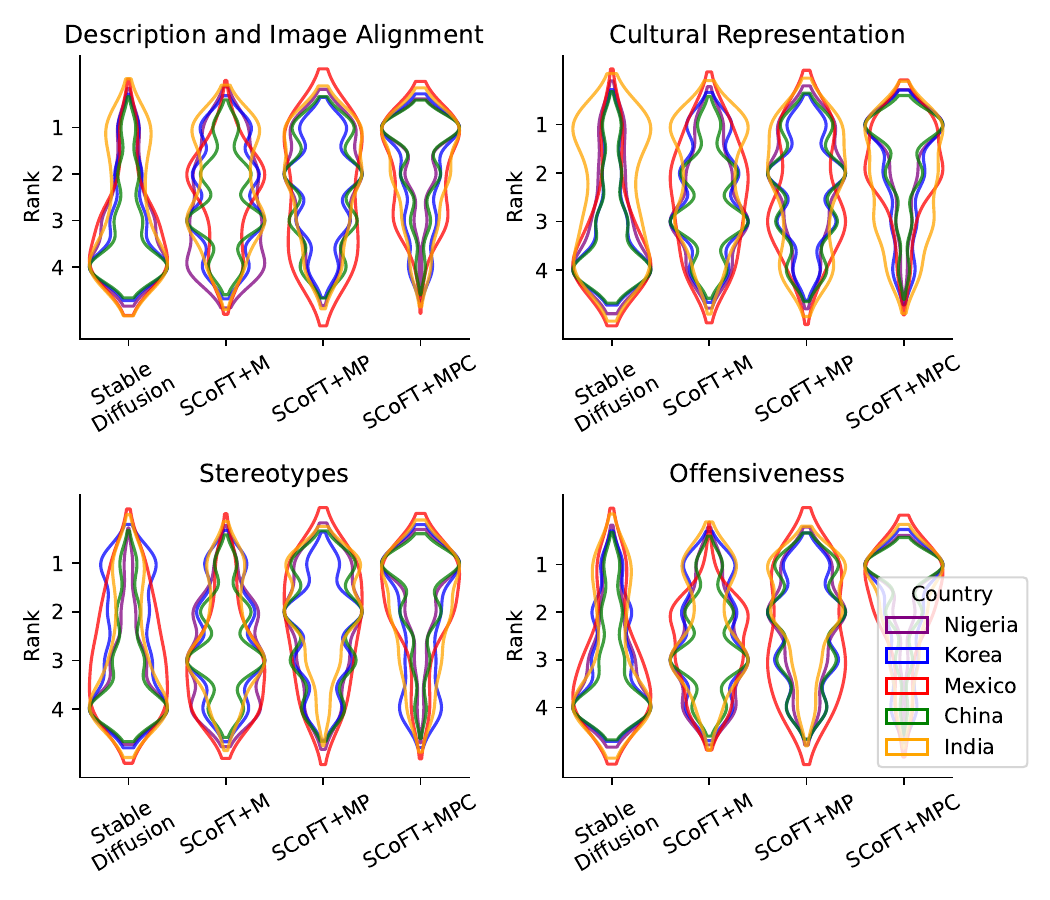}
    \caption{Violin plot of participant rankings across the survey items and countries. A wider strip means more answers with that value. Each new loss in our ablation study improved the rankings, and SCoFT+MPC is best. (Rank 1 is the best; 4, the worst)}
    \label{fig:violin}
    \vspace{-0.5cm}
\end{figure}

We compare whole distributions of the rankings in Figure~\ref{fig:violin}. Across survey items, we see very similar distributions. For example, Stable Diffusion images were very commonly ranked fourth for both Stereotypes and Cultural Representation.
Participants in the Chinese and Korean surveys ranked images with less variance than participants in the Indian and Mexican surveys. This is potentially due to a difference in the number of participants for each survey.

\noindent\textbf{Automatic Metric Results.} \label{sec:automatic_results}
%
%
The automated evaluation results are summarized in Table~\ref{tab:survey} where the proposed approach achieves the best result in terms of the KID score based on the CCUB test set, indicating that the fine-tuned model is able to generate images with a similar quality to the culturally curated data. 
By contrast, the original Stable Diffusion model achieves the highest score in terms of the CLIP Score that measures the text-image alignment as well as the KID score based on the MS COCO dataset. 
This result is not surprising as the CLIP model is itself known to be biased~\cite{birhane2021stereotypesInLAION} in ways shared with Stable Diffusion~\cite{luccioni2023stable}, where the number of measurable outputs people perceive as hatred scales with the training set~\cite{birhane2023hate}. 
CLIP biases have also been quantitatively shown to be passed on to downstream applications~\cite{hundt2022robots_enact}.
Since human evaluators found our SCoFT+MPC method outperforms Stable Diffusion, we conclude that CLIP-Score is not an appropriate metric for assessing cultural competence.

\noindent\textbf{Perceptual Backbones and Gradient Recording.}\label{sec:perceptual_and_gradient}
SCoFT uses a perceptual backbone to compare images in feature spaces rather than pixel space to avoid overfitting to training images.
We test several backbones to extract image features, including the output of CLIP~\cite{radford2021learning} convolutional layers~\cite{vinker2022clipasso}, and the last layer of DreamSim~\cite{fu2023dreamsim}, DINOv2 \cite{oquab2023dinov2}, BLIP2 \cite{li2023blip}. For each comparison, we generate 200 images and calculate the KID with CCUB test set with the Korean, Chinese, and Indian models and see that the CLIP convolutional layers and DreamSim output provide the best generalization. For all other experiments, we use the CLIP convolutional layers as the SCoFT backbone.

\begin{figure}[t]
  \centering \small
   \includegraphics[width=0.95\linewidth]{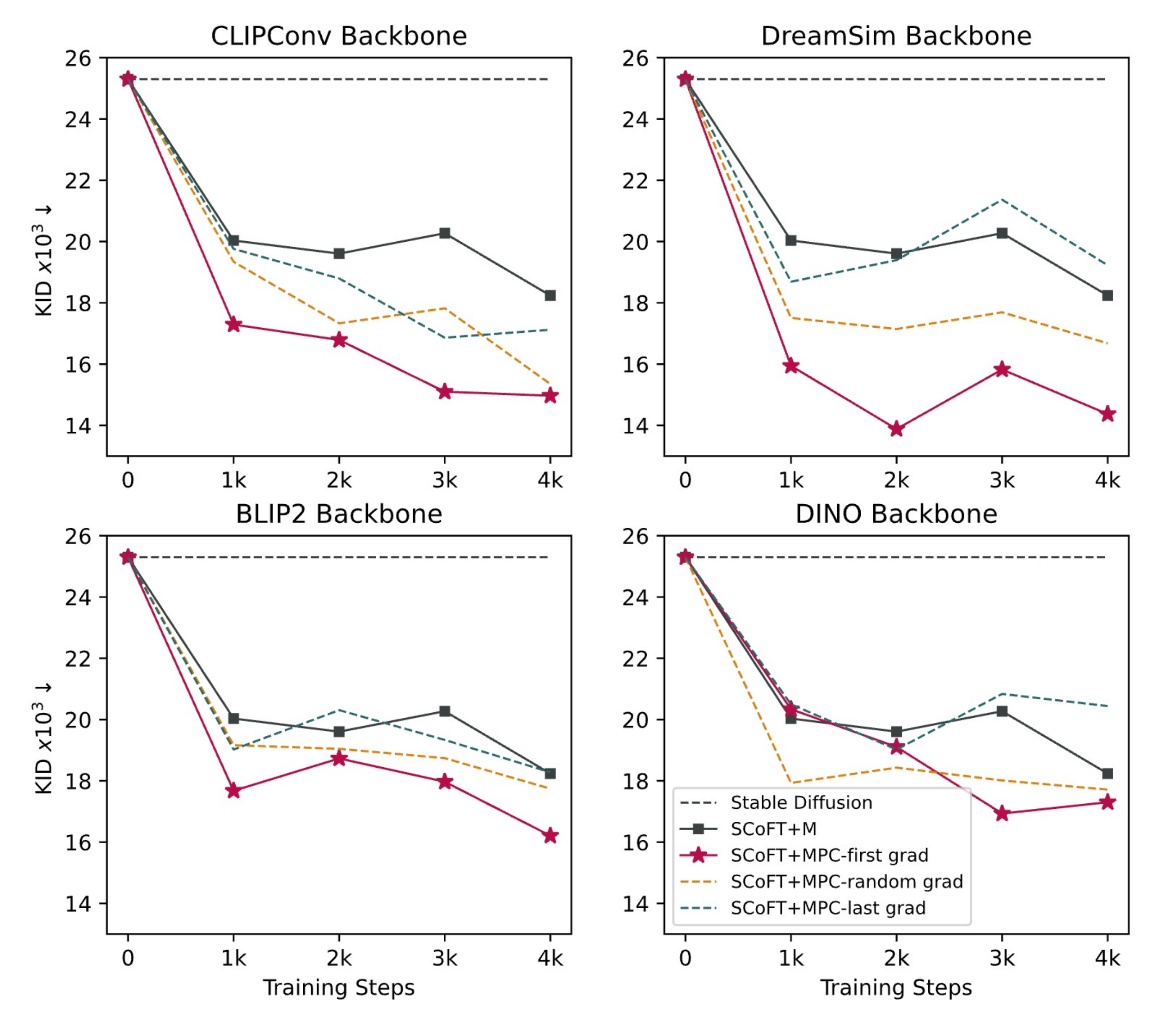}\vspace{-10pt}%
   \caption{Comparison of perceptual models used for feature extraction in SCoFT. KID between held out CCUB images and generated images is plotted versus training iterations representing generalization to a validation set, score averaged across three cultures.}%
   \label{fig:backbone}%
   \vspace{-0.3cm}%
\end{figure}

We also compare the effect of recording the gradient during the first, last, and random iterations of denoising during fine-tuning, as reported in Figure~\ref{fig:backbone}. We see that the best generalization comes when recording the gradient during the first iteration of denoising. 
This is in contrast to concurrent work \cite{clark2023directly} which recorded the last gradient, indicating that culture is a high-level concept where important information is created early in the diffusion process, as opposed to aesthetics which are low-level and correspond more strongly to later stages of diffusion.

\begin{table*}
  \centering
  \small
  \begin{tabular}{@{}l||cccc||cc@{}}
    \toprule
         \multicolumn{1}{c||}{} &
     \multicolumn{4}{c||}{Prompts from Training set} &
     \multicolumn{2}{c}{Prompts from Test set} 
     \\
    \hline
    Method & CLIP-I $\downarrow$ & DINO $\downarrow$ & DreamSim $\downarrow$ & DIV train $\uparrow$ & DIV test $\uparrow$ & CLIPScore $\uparrow$\\
    \midrule
    Finetune   & 0.912 & 0.836& 0.591 & 0.302 & 0.317& \textbf{0.824} \\
    Finetune w/ $\mathcal{L}_{M}$ & \textbf{0.897} & \textbf{0.808}& \textbf{0.550} & \textbf{0.356} & \textbf{0.379}& 0.814\\
    \bottomrule
  \end{tabular}
  \vspace{-0.2cm}
  \caption{
  Images generated from Stable Diffusion fine-tuned with and without Memorization Loss, $\mathcal{L}_{M}$,  are compared in the feature space from various feature extractors. We see that $\mathcal{L}_{M}$ encourages the model to produce images with more diversity (larger feature difference).
  }
  \label{tab:memorization}
  \vspace{-0.5cm}
\end{table*}

\begin{figure}[t]
  \centering%
   \includegraphics[width=0.9\linewidth]{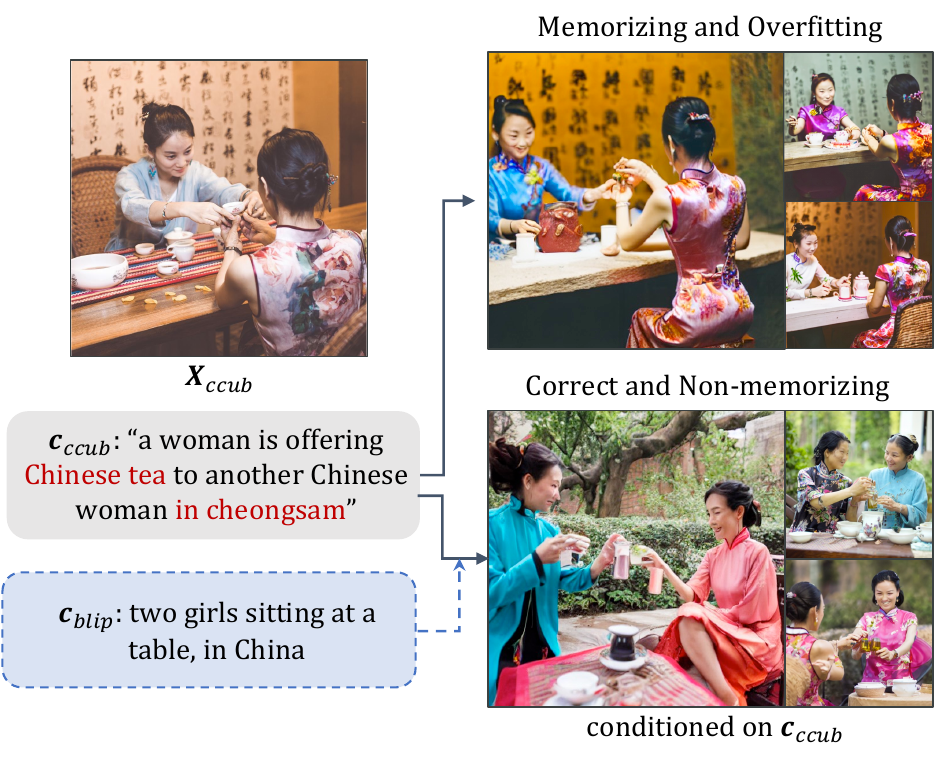}\vspace{-10pt}%
   \caption{
   Top-right: Fine-tuning Stable Diffusion on CCUB data using only a conventional loss ($\mathcal{L}_{LDM}$, Sec.~\ref{subsec:background}) leads to overfitting on CCUB captions. Bottom-right: Adding memorization loss ($\mathcal{L}_{M}$, Sec.~\ref{subsec:memorization_loss}) prevents overfitting with small datasets by ensuring images generated by general captions ($\mathbf{c}_{blip}$) are similar to those generated using CCUB's cultural captions. 
   }%
   \label{fig:overfitting}\vspace{-5pt}%
\vspace{-0.4cm}%
\end{figure}%
%

\noindent\textbf{Effectiveness of the Memorization Loss.}
%
We fine-tune Stable Diffusion using the original $\mathcal{L}_{LDM}$ with and without Memorization Loss, $\mathcal{L}_M$, on the CCUB dataset.
To evaluate the consequence of over-fitting and reproducing training images during few-shot fine-tuning, we randomly select 10 text-image pairs for each culture from the CCUB training set. For each training text prompt, we generate 20 images.
We evaluate the process using three metrics: CLIP-Image (CLIP-I), DINO, and DreamSim. All metrics measure the average pairwise cosine similarity between the embeddings of generated and real training images that share the same text prompt. 
For both metrics, lower values signify more effective prevention of overfitting and reproduction of training images. Figure~\ref{fig:overfitting} shows the qualitative results of generating more creative images. Quantitative results in Table~\ref{tab:memorization} show that the memorization loss effectively reduces overfitting.

To quantify output diversity, we randomly select 10 training text prompts and 10 CCUB testing text prompts. For each text prompt, we generate 20 images. We introduce the diversity metric (DIV), which calculates the average pair-wise DreamSim cosine distance between generated images with same text prompt. Higher values indicate enhanced diversity in the generated outputs, reflecting a more varied and expressive synthesis of images.
We also report a comparable CLIP Score on the generated image using CCUB testing text prompts with baseline fine-tuning.


\begin{figure}
    \centering
    \includegraphics[width=1\linewidth]{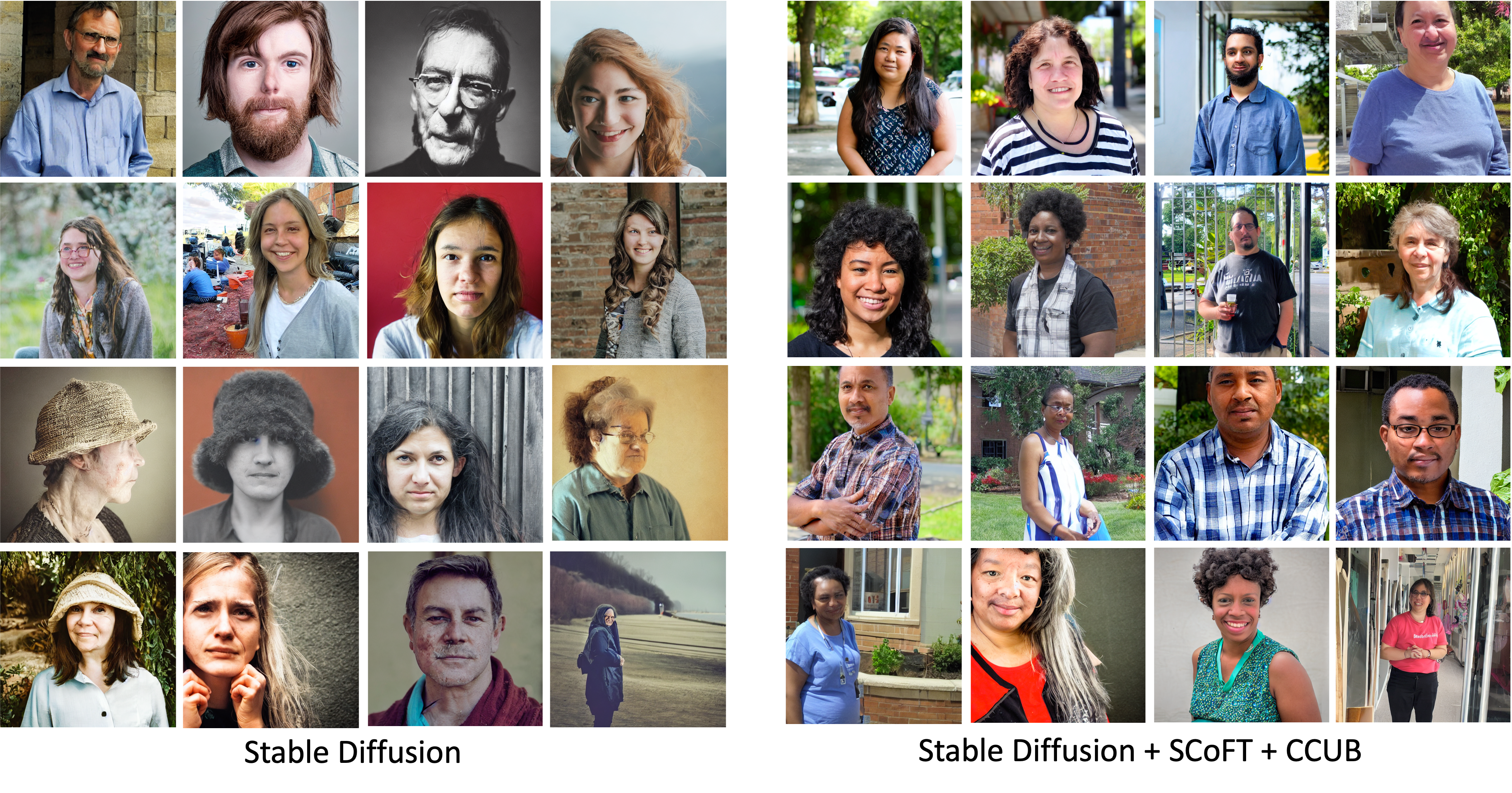}\vspace{-10pt}%
    \caption{Generated images for text prompt: ``a photo of a person.'' The proposed approach is able to generate more diverse images given a generic prompt without a specific cultural context.}
    \label{fig:diversity}%
    \vspace{-0.5cm}
\end{figure}%

%% file: sec/8_conclusion.tex
\noindent\textbf{Limitations.}
To tackle the bias in the data, we aim for two goals: 1) to generate accurate images given a specific cultural context and 2) to generate diverse images given a generic text prompt without any specific cultural context. Our current approach is focused on achieving the first goal. Our current model can generate promisingly diverse images for some generic prompts as shown in Figure~\ref{fig:diversity} when compared to the baseline model that generates biased images.  

Our CCUB dataset was collected by experienced residents; however, to improve the quality of the dataset, more vigorous verification will be needed. 

%
\vspace{-0.2cm}
\section{Conclusion}
\vspace{-0.2cm}
The biases of generative AI have already led to substantial and very public impacts~\cite{fourzeromedia2023marketplace,Lensa2022}, so it is essential that we ensure models generate images that more accurately represent the diversity of the world. 
We propose Self-Contrastive Fine-Tuning (SCoFT), which is specifically designed to fine-tune the model for high-level concepts using a small dataset, for instance, to pay attention to subtle cultural elements.
Our proposed methods have potential applications in other domains such as reducing the risk of copyright infringement, better respecting cultural and community-defined boundaries, and addressing offensiveness across a broader range of identity characteristics and other criteria. 
We confirm positive associations with some automated metrics while demonstrating others are not a good fit for this task.
 
Our extensive user survey and metric evaluation quantitatively demonstrate improvements in subjective metrics with respect to image and description alignment, more culturally representative image contents, as well as reductions in stereotyping and offensiveness. 

%% file: sec/9_acknowledgements.tex
\noindent\textbf{Acknowledgment.} 
We are grateful to Lia Coleman, Youeun Shin for their collaboration, to Chaitanya Chawla, Pablo Ortega, Vihaan Misra, Ingrid Navarro, Yu Chen, Jinyun Xu, Jiaji Li, Mingzhu Liu for contributing the CCUB dataset collection, and to Nupur Kumari, Zeyu Qin, Heng Yu, Jinqi Luo for their helpful discussion.
This work was in part supported by NSF IIS-2112633 and the Technology Innovation Program (20018295, Meta-human: a virtual cooperation platform for specialized industrial services) funded By the Ministry of Trade, Industry \& Energy(MOTIE, Korea).
Youngsik Yun, and Jihie Kim were supported by the MSIT (Ministry of Science, ICT), Korea, under the High-Potential Individuals Global Training Progracm (RS-2022-00155054) supervised by the IITP (Institute for Information \& Communications Technology Planning \& Evaluation) (50\%). 
Andrew Hundt's contributions are based upon work supported by the National Science Foundation under Grant \# 2030859 to the Computing Research Association for the CIFellows Project with subaward \# 2021CIF-CarnegieMellon-72.

%% file: sec/X_suppl.tex
\clearpage
\setcounter{page}{1}
\maketitlesupplementary
\noindent \textbf{Overview.} In Sec.~\ref{sec:app_SCoFT}, we present the pseudocode for SCoFT and report the training details for the experiments. Sec.~\ref{sec:app_CCUB_dataset} provides statistics about the CCUB dataset. Sec.~\ref{sec:app_automatic} showcases the results of automatic metrics per culture, followed by the human evaluation survey question and additional analysis on human evaluation feedback in Sec.~\ref{sec:AMT}. Finally, Sec.~\ref{sec:app_qualitative} includes more examples for our models, and Sec.~\ref{sec:app_limitation} discusses the ethics and limitations.

\section{SCoFT Method}
\label{sec:app_SCoFT}
\subsection{SCoFT Pseudocode}
We show here the pseudocode of our SCoFT method.

\begin{algorithm}
    \caption{SCoFT for Stable Diffusion}
    \label{scoft_pseudocode} 
    \begin{algorithmic}
    \State \textbf{dataset:} CCUB dataset $\mathcal{D}_{ccub} = \{(\mathbf{x}_{ccub}, \mathbf{c}_{ccub}, \mathbf{c}_{blip})\}$ 
    \State \textbf{inputs:} Pre-trained Stable Diffusion 1.4 model $\epsilon_{\theta}$, LoRA weights $\hat{\theta}$
    \For{$(\mathbf{x}^i_{ccub}, \mathbf{c}^i_{ccub}, \mathbf{c}^i_{blip}) \in \mathcal{D}_{ccub}$}
    \State $\textbf{z}_0 = ENCODE(\mathbf{x}^i_{ccub})$
    \State $\epsilon_t \sim \mathcal{N}(0,\mathbf{I})$
    \State $\mathbf{z}_t = \sqrt{\alpha_t}\mathbf{z}_0 + \sqrt{1-\alpha_t}\epsilon_t$
    \State $\mathcal{L}_{LDM} = MSE(\epsilon_t, \epsilon_{\hat{\theta}}(\textbf{z}_0, \textbf{c}_{ccub},t))$ \Comment{$\mathcal{L}_{LDM}$}
    \State $\mathcal{L}_M = MSE(\epsilon_{\hat{\theta}}(\textbf{z}_0, \textbf{c}_{ccub},t), \epsilon_{\hat{\theta}}(\textbf{z}_0, \textbf{c}_{blip},t))$ \Comment{$\mathcal{L}_{M}$}
    \If{every 10 timesteps}
        \State $\mathbf{x}^+ = \mathbf{x}_{ccub}$
        \State $\mathbf{x}^- = \{ \Theta(\mathcal{D}_{depth}(\mathbf{x}^+),\mathbf{c}_{blip})\}$
        \State $t_{record} = \left\{ \begin{array}{rcl}
            t       &      & \text{record first gradient}\\
            t_{rand}     &      & \text{record random gradient}\\
            1    &      & \text{record last gradient}
            \end{array} \right.$
            \State $\mathbf{\hat{z}}_t = \mathbf{z}_{t}$
            \For{$u = t, ..., 1$}
                \If{$u \neq t_{record}$}
                    \State $\mathbf{\hat{z}}_{u} = \text{stop\_grad}(\mathbf{\hat{z}}_{u})$
                \EndIf
                \State $\mathbf{\hat{z}}_{u - 1} = \epsilon_{\hat{\theta}}(\mathbf{\hat{z}}_{u}, \textbf{c}_{ccub},u)$  // de-noise
            \EndFor
        \State $\hat{\mathbf{x}} = DECODE(\hat{\mathbf{z}}_0)$
        \State $\mathcal{L}_{C}(\hat{x}, x^+, x^-)
            = \mathbb{E}_{\hat{x}, x^+, x^-}[\max(\mathcal{S}(\hat{x}, {x}^+;f_{\theta})$
        \State $-\lambda \mathcal{S}(\hat{x}, {x}^-;f_{\theta}) + m, 0)]$ \Comment{$\mathcal{L}_{P} + \mathcal{L}_{C}$}
    \EndIf
    \State $\mathcal{L} = \lambda_{l}\mathcal{L}_{LDM} + \lambda_{m}\mathcal{L}_{M} + \lambda_{c}\mathcal{L}_{C}$
    \State $g = \nabla \mathcal{L}(\theta^{LoRA})$
    \State $\theta^{LoRA} \gets \theta^{LoRA} - \eta g$
    \EndFor
    \end{algorithmic}
\end{algorithm}

\subsection{Training Details}
We detail the training specifics and hyperparameters and maintain uniform settings for each model for a fair comparison. Throughout training, we employ the Adam optimizer with $\beta_1 = 0.9$ and $\beta_2 = 0.999$ for 3000 iterations, utilizing a learning rate of $1e-4$. The batch size is set to 1, and LoRA is exclusively applied to UNet parameters with a rank of 64.
We select CLIPConv using the fifth convolutional layer as the backbone and record the first gradient during backpropagation through sampling. For each positive example, we generate 5 negative examples, employing DreamSim to filter out false negatives which are similar to positive sample. As the training weights for different losses,, we set $\lambda_{l} = 0.7$ and $\lambda_{m} = 0.3$. To manage time costs, we compute $\mathcal{L}_{c}$ every 10 iterations, denoising the latent $\mathbf{z}_t$ using 20 steps of DDIM sampling, with $\lambda_{c} = 0.1$. The training process runs for approximately 2.5 hours on a single NVIDIA V100 GPU for each CCUB cultural dataset.
\vspace{-0.1cm}
\section{CCUB Dataset Scale}
\label{sec:app_CCUB_dataset}
Table \ref{tab:cultural-datasets} illustrates the scale of our CCUB dataset. The
number of hand-selected images and their corresponding captions
in nine cultural categories for six different cultures are listed.

CCUB contains the image web links and the manual captions. We do not own the copyright of the images. 

\begin{table}[]
\centering
\resizebox{\columnwidth}{!}{%
\begin{tabular}{|l||lllllllll||l|}
\hline
 &
  \rotatebox{90}{food \& drink} &
  \rotatebox{90}{people \& actions} &
  \rotatebox{90}{clothing} &
  \rotatebox{90}{architecture} &
  \rotatebox{90}{city} &
  \rotatebox{90}{dance music art } &
  \rotatebox{90}{nature} &
  \rotatebox{90}{utensil \& tool} &
  \rotatebox{90}{religion \& festival} &
  \rotatebox{90}{\textbf{total}}\\ \hline 
Korea & 34 & 22 & 20 & 20 & 21 & 20 & 8 & 6 & 11 & 162 \\ \hline 
China & 33 & 31 & 24 & 23 & 27 & 23 & 5 & 15 & 8 & 189 \\ \hline
Nigeria & 21 & 16 & 18 & 17 & 13 & 21 & 11 & 8 & 15 & 140 \\ \hline
Mexico & 22 & 19 & 14 & 18 & 15 & 10 & 7 & 10 & 19 & 134 \\ \hline
India & 19 & 26 & 24 & 21 & 16 & 18 & 9 & 7 & 8 & 148 \\ \hline
\begin{tabular}[c]{@{}l@{}}United\\  States\end{tabular} & 23 & 22 & 10 & 17 &29 & 15 & 16 & 12 & 7 & 151 \\ \hline
\textbf{Total} & 152 & 136 & 110 & 116 & 121 & 107 & 56 & 58 & 68 & \textbf{924} \\ \hline

\end{tabular}%
}
\caption{
This table shows the scale of our CCUB dataset, detailing the number of hand-selected images and their corresponding captions across nine cultural categories for six different cultures.}
\label{tab:cultural-datasets}
\vspace{-0.5cm}
\end{table}
\section{Automatic Metrics}
\begin{figure}
    \centering
    \includegraphics[width=\columnwidth]{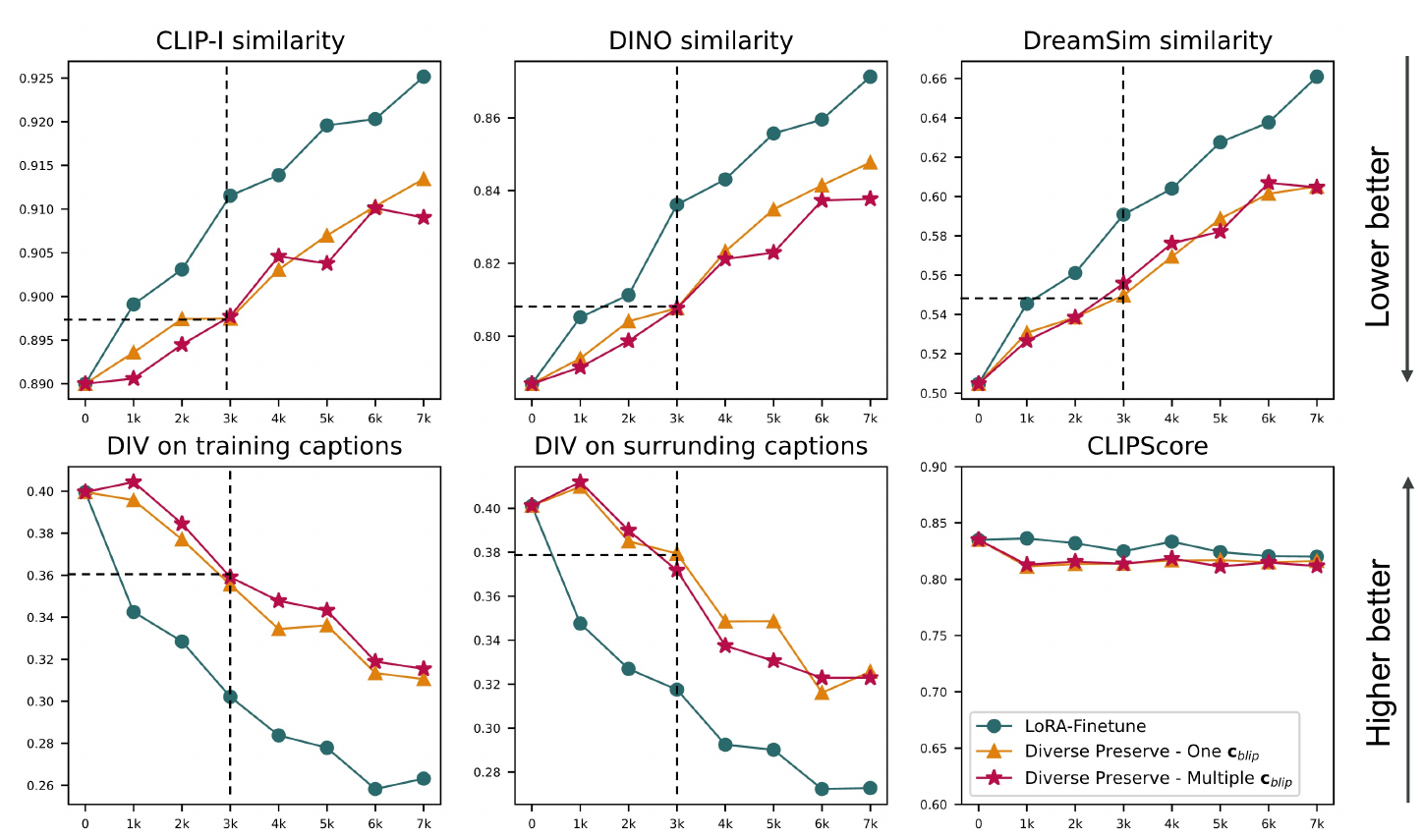}
    \caption{\textbf{Preventing Memorizing during training}. We compare training with (``LoRA-Finetune") and without $\mathcal{L}_{M}$ (``Diverse Preserve"), analyzing the effects of using one   or multiple $\mathbf{c}_{blip}$. The scores are averaged across three CCUB culture datasets. Fine-tuning with $\mathcal{L}_{M}$ for 3000 steps achieves even better performance than without it for 1000 steps.
    Adding $\mathcal{L}_{M}$ effectively prevents overfitting, and we notice that using more than one $\mathbf{c}_{blip}$ has little effect.
    }
    \label{fig:app_diverse_metric}
\end{figure}

Figure \ref{fig:app_diverse_metric} demonstrates that $\mathcal{L}_M$ serves to prevent memorization and enhance diverse expression. The metrics are calculated during training and averaged across three cultures. Additionally, Table \ref{tab:app_automatic_metrics} offers supplementary details, abalating SCoFT for automatic metrics performance across each culture.

\label{sec:app_automatic}
\begin{table*}[h!]
    \scriptsize
    \centering
    \begin{tabular}{l||ccccc||ccccc||ccccc}
    \toprule
    \multicolumn{1}{c||}{} &
     \multicolumn{5}{c||}{KID-CCUB $\times 10^3 \downarrow$} &
     \multicolumn{5}{c||}{KID-COCO$\times 10^3 \downarrow$} &
     \multicolumn{5}{c}{CLIP Score $\uparrow$} 
     \\
    \hline
    Model Name 
    & CN
    & KR
    & NG
    & IN
    & MX
    & CN
    & KR
    & NG
    & IN
    & MX
    & CN
    & KR  
    & NG
    & IN
    & MX \\ \midrule
    Stable Diffusion 
    & 25.244& 28.061& 32.687 & 39.988 & 25.796 
    & \textbf{4.396} &  4.396 &  \textbf{4.396} & \textbf{4.396} & 4.396  
    &\textbf{0.827} & \textbf{0.844} & \textbf{0.799} & \textbf{0.773} & 0.819 \\
    SCoFT+M     
    & 18.798  &24.118  &24.539 & 20.566 & 25.196 
    & 4.890& 4.497 & 4.461 & 5.342 & 4.364 
    &0.804 & 0.832 & 0.792 & 0.757 & 0.825\\
    SCoFT+MP    
    &15.895  & 23.007 &25.638  &19.799 &\textbf{22.462}
    &5.090 & 4.403& 5.135&5.112&4.941
    &0.798 &0.836 &0.792 &0.753 &0.8232\\
    SCoFT+MPC   
    &\textbf{14.667} &\textbf{17.413} &\textbf{23.261} &\textbf{18.342} &24.424
    &5.102 & \textbf{4.222} &4.987&5.457 &\textbf{4.325}
    &0.793 &0.825 &0.795 &0.753 &\textbf{0.829}\\
    \bottomrule
    \end{tabular}
    \caption{We compare our SCoFT ablations to Stable Diffusion using automatic metrics and present the results for five cultures. The KID-CCUB is calculated on the internal CCUB test dataset, containing 150 cultural data for each culture. KID-COCO is calculated on randomly selected 500 text-image pairs from the MS-COCO dataset. CLIPScore is calculated using the CCUB test dataset text prompts.}
    \label{tab:app_automatic_metrics}
\end{table*}

\section{Human Evaluation}
\label{sec:AMT}

\begin{figure*}[t]
    \centering
    \begin{subfigure}[t]{0.3\textwidth}
        \centering
        \includegraphics[width=0.9\linewidth]{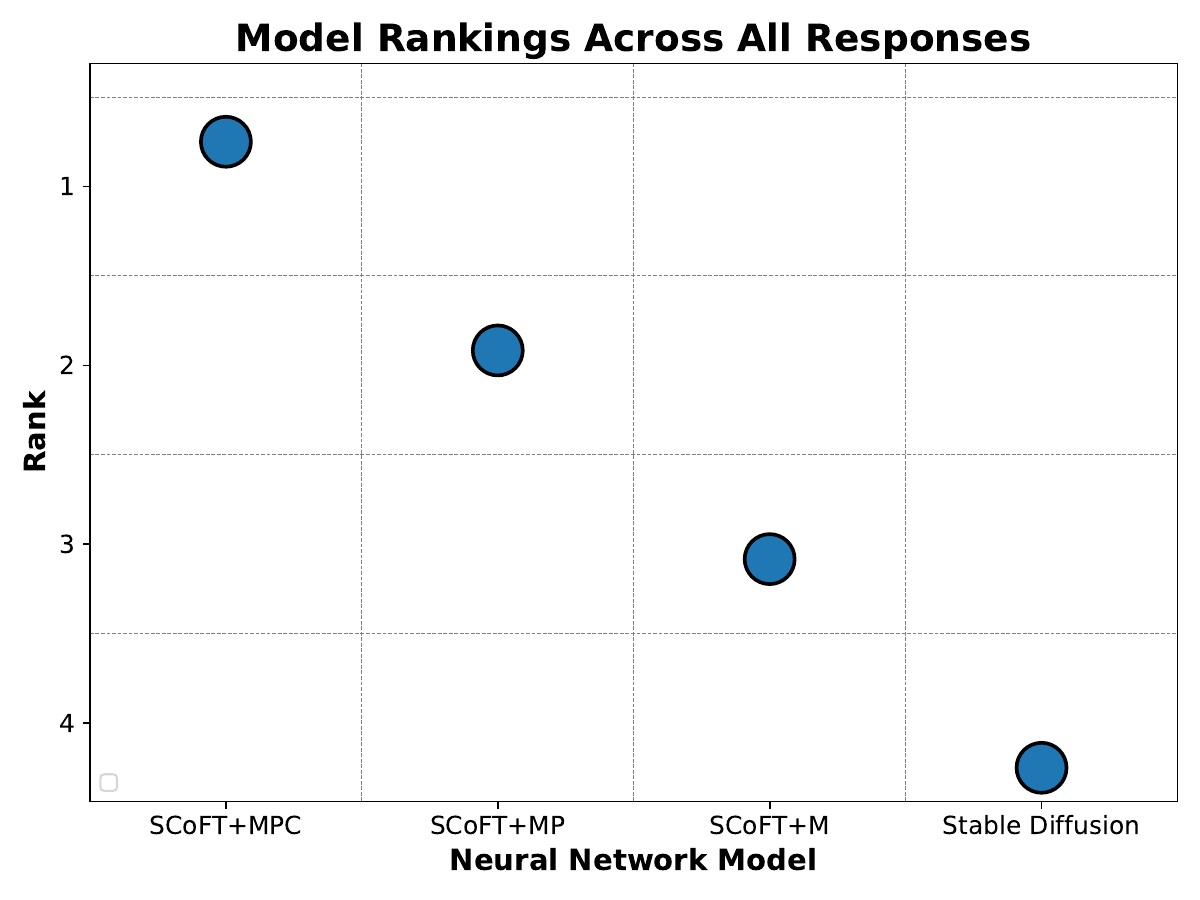}
        \caption{Overall, participants find our SCoFT+MPC method outperforms all comparable methods when ranked with MMSR+Vote.}
        \label{fig:mmsr_overall}
    \end{subfigure}
    \hfill
    \begin{subfigure}[t]{0.3\textwidth}
        \centering
        \includegraphics[width=0.9\linewidth]{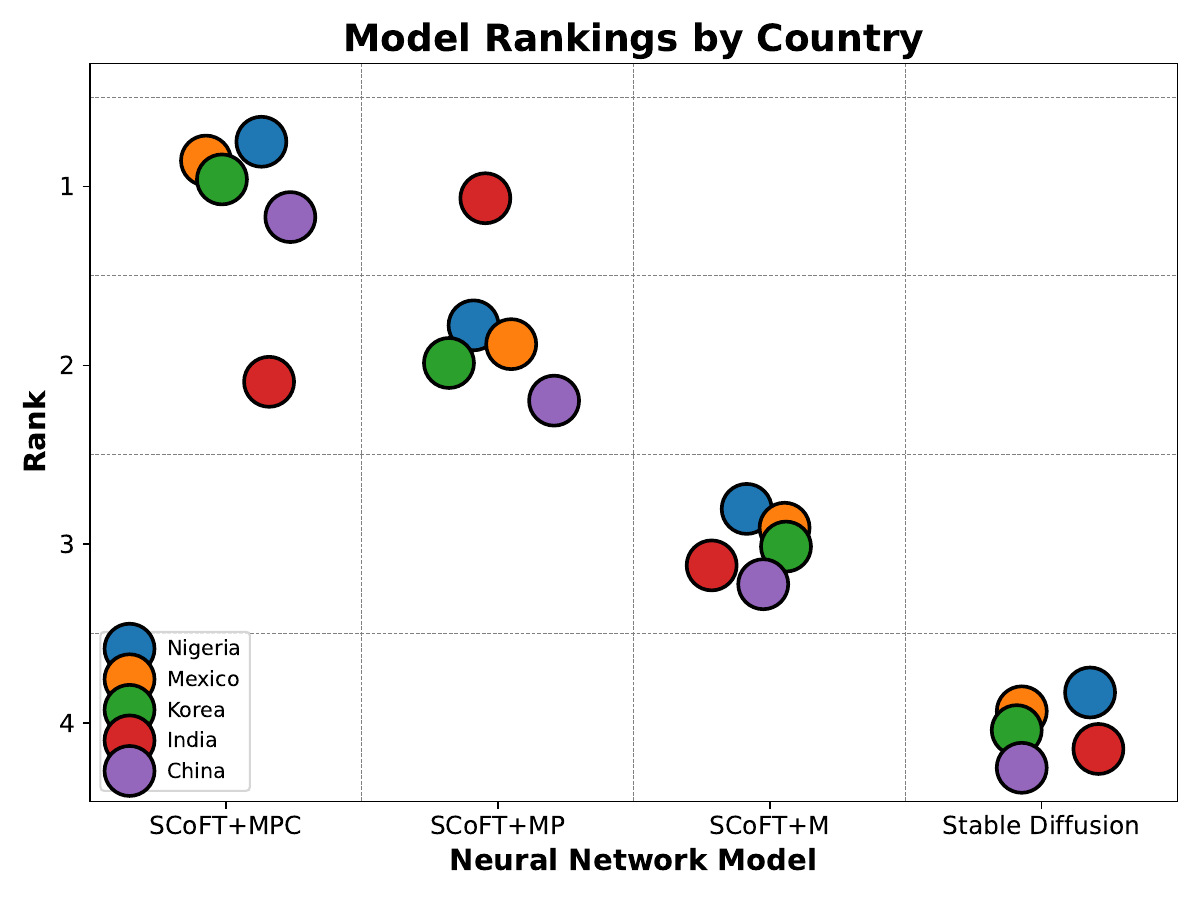}
        \caption{Participants' overall preference for our SCoFT+MPC method is generally consistent across national affiliations when ranked with MMSR+Vote.}
        \label{fig:mmsr_country}
    \end{subfigure}
    \hfill
    \begin{subfigure}[t]{0.3\textwidth}
        \centering
        \includegraphics[width=0.9\linewidth]{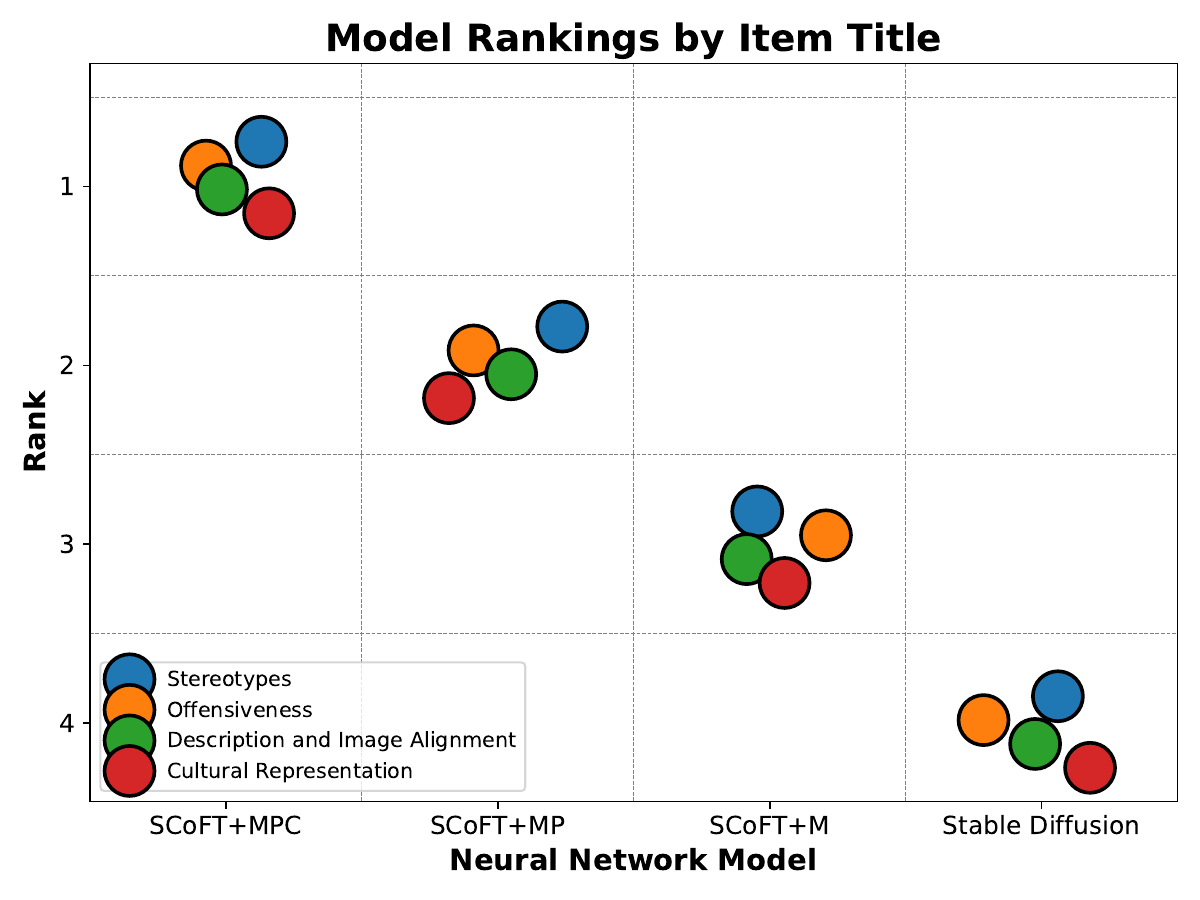}
        \caption{Participants prefer our SCoFT+MPC method with respect to Description and Image Alignment, Cultural Representation, Stereotyping, and Offensiveness when ranked with MMSR+Vote.}
        \label{fig:mmsr_survey_item}
    \end{subfigure}
    \caption{Methods Ranked According to Culturally Experienced Human Participants (MMSR+Vote alg, see Sec. \ref{subsubsec:mmsr_vote}) Higher ranks (numerically smaller values) are better.}
    \label{fig:mmsr_combined}
\end{figure*}

\begin{figure*}[t]
    \centering
    \begin{subfigure}[t]{0.3\textwidth}
        \centering
        \includegraphics[width=0.9\linewidth]{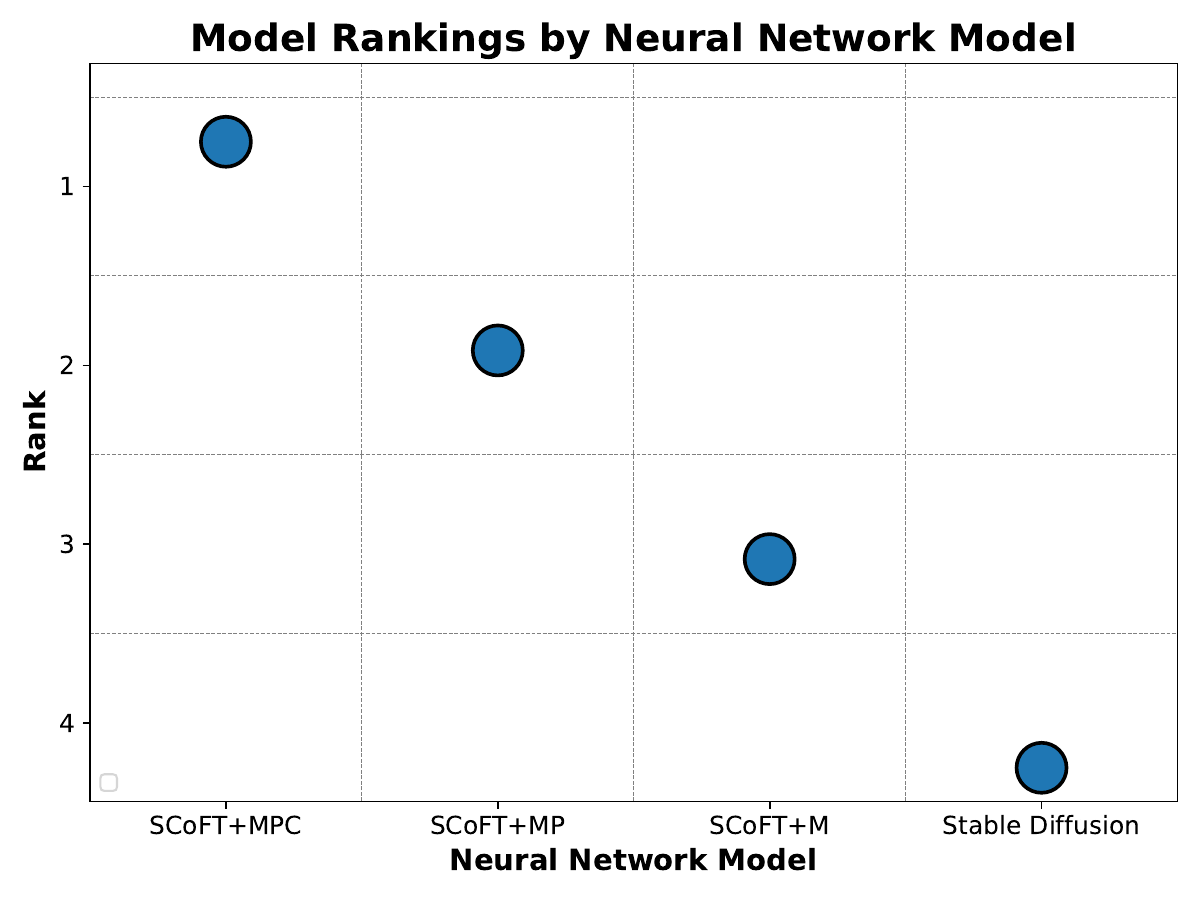}
        \caption{Overall, participants find our SCoFT+MPC method outperforms all comparable methods when ranked with the Noisy Bradley Terry Algorithm.}
        \label{fig:nbt_overall}
    \end{subfigure}
    \hfill
    \begin{subfigure}[t]{0.3\textwidth}
        \centering
        \includegraphics[width=0.9\linewidth]{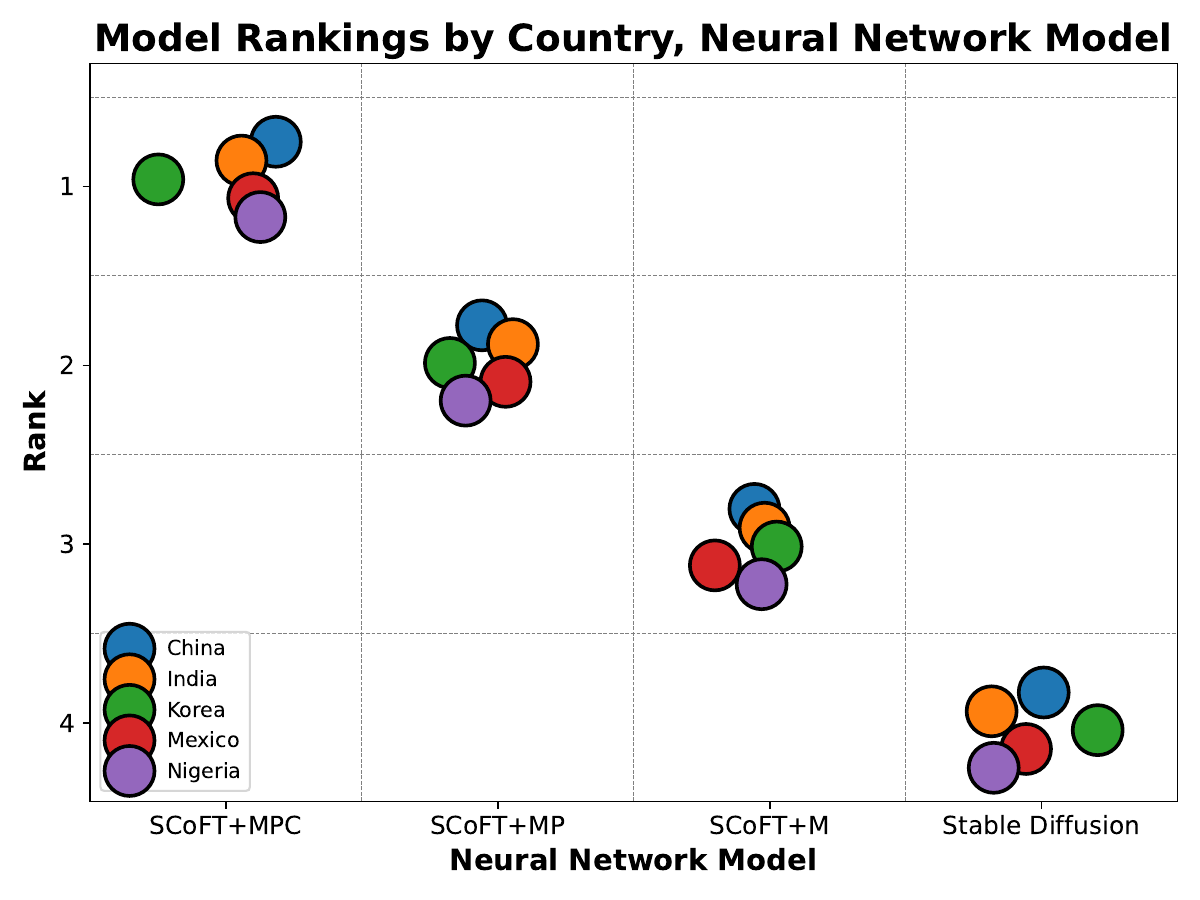}
        \caption{Participants' overall preference for our SCoFT+MPC method is consistent across national affiliations when ranked with the Noisy Bradley Terry Algorithm.}
        \label{fig:nbt_country}
    \end{subfigure}
    \hfill
    \begin{subfigure}[t]{0.3\textwidth}
        \centering
        \includegraphics[width=0.9\linewidth]{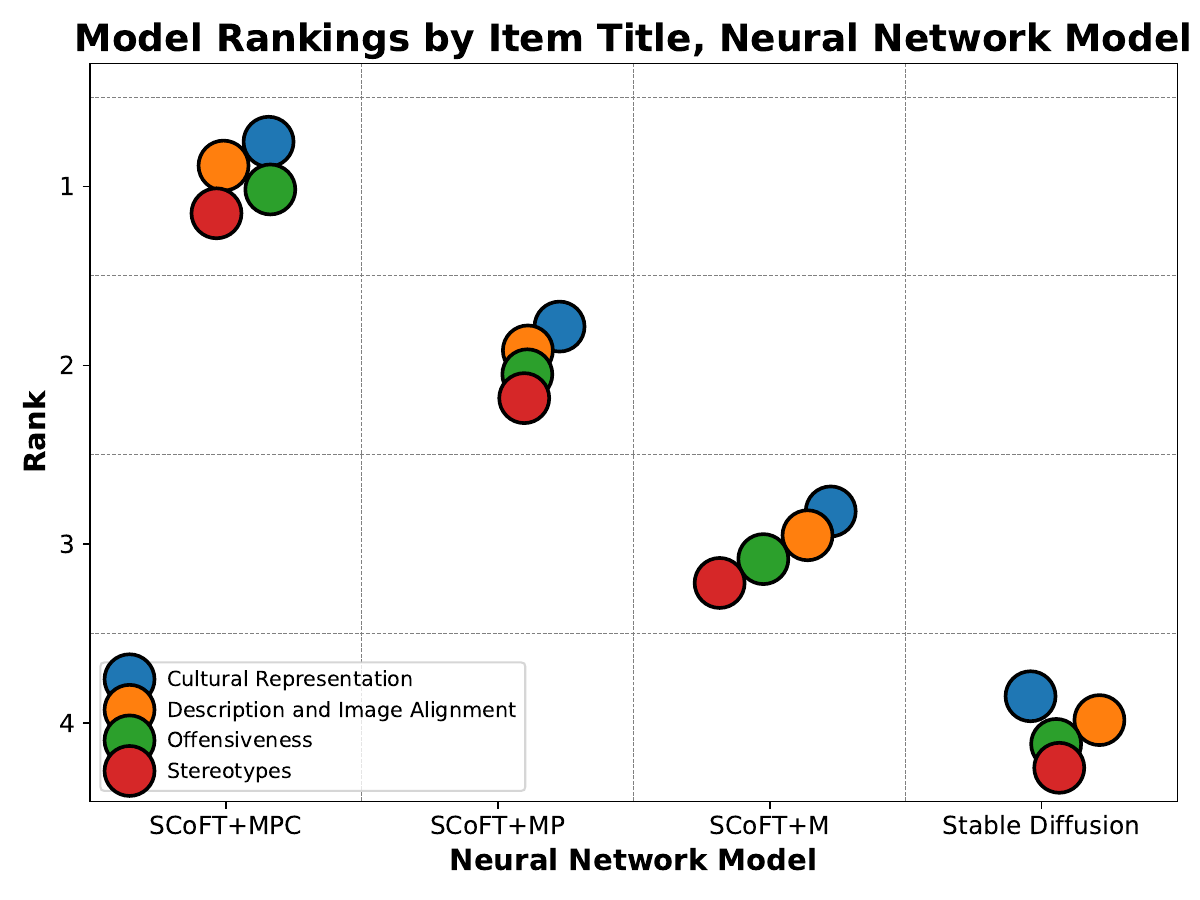}
        \caption{Participants prefer our SCoFT+MPC method with respect to Description and Image Alignment, Cultural Representation, Stereotyping, and Offensiveness when ranked with the Noisy Bradley Terry Algorithm.}
        \label{fig:nbt_survey_item}
    \end{subfigure}
    \caption{Methods Ranked According to Culturally Experienced Human Participants (Noisy Bradley Terry alg, see Sec. \ref{subsubsec:mmsr_vote}) Higher ranks (numerically smaller values) are better.}
    \label{fig:nbt_combined}
\end{figure*}

\begin{table*}[!htbp]
  \centering
  \begin{tabular}{@{}lp{10cm}@{}}
    \toprule
    \textbf{Item Title} & \textbf{Item Text} \\
    \midrule
    Description and Image Alignment & Rank the accuracy of the match between the image and its description. Number each image from 1 for the image best described by the description to 4 for the worst match between image and description. \textbf{(1=best described, 4=worst described)} \\
    \addlinespace
    Cultural Representation & Rank the images from 1 for the best representation of ${country\_adj}$ culture to 4 for the worst cultural representation. Please ignore image artifacts (such as distorted faces, hands, or glitches) when considering Cultural Representation. \textbf{(1=most representative, 4=least representative)} \\
    \addlinespace
    Stereotypes & Rank the images from 1 for being the least stereotypical depiction of ${country\_adj}$ culture to 4 for the most based on your personal knowledge of ${country\_adj}$ stereotypes. \textbf{(1=least stereotypical, 4=most stereotypical)} \\
    \addlinespace
    Offensiveness & Rank the images by their offensiveness to you personally, numbering from 1 as the least offensive to 4 for the most offensive. Please ignore image artifacts (such as distorted faces, hands, or glitches) when considering Offensiveness. \textbf{(1=least offensive, 4=most offensive)} \\
    \bottomrule
  \end{tabular}
  \caption{The four survey items for our human evaluation survey comparing images generated using different model variations.}
  \label{tab:survey_items}
\end{table*}





\subsection{Survey Description}
\label{subsec:amt_appendix}

\begin{figure*}
    \centering
    \includegraphics[width=0.65\textwidth]{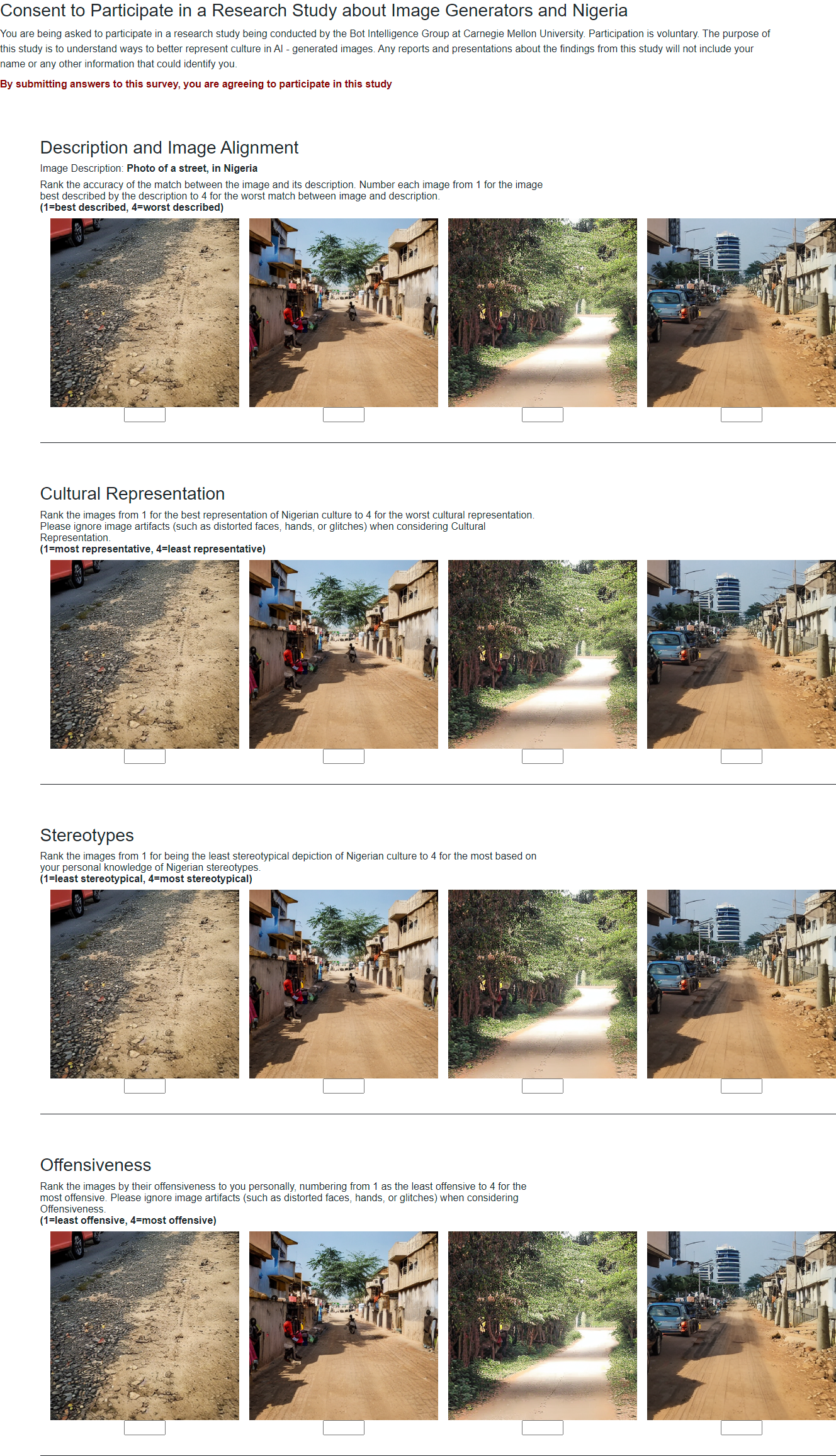}
    \caption{A sample of a single survey page. Participants enter a rank from 1 to 4 in the white text boxes immediately below each image to indicate their perspective. Four values are entered for each survey item: Description and Image Alignment, Cultural Representation, Stereotypes, Offensiveness. Each set of four images is in one consistent randomized order for that page.}
    \label{fig:survey_page}
\end{figure*}

Our goal of improving the cultural perception of generated images is a subjective metric largely determined by members of a given identity group. 
To evaluate our performance on this criteria, we recruited people with at least 5 years of cultural experience in each of the 5 countries with survey questions specific to their self-selected national cultural affiliation. 
A single page of the survey form provides one description (prompt) and one image made by each of the four generators using a common random seed, for four total images.
Each survey page has a total of four survey items (rows that participants respond to, see Table \ref{tab:survey_items}) to rank relative to (a) Description and Image Alignment, (b) Cultural Representation, (c) Stereotypes, and (d) Offensiveness.
Participants respond by numerically ranking the set of randomly ordered images from best image to worst image once for each item.
An image labeled rank 1 would signify both best aligned and least offensive when each case is ranked, while rank four would be least well aligned and most offensive.
A sample of a single survey page can be viewed in Figure \ref{fig:survey_page}. 

\subsection{Analysis and Evaluation of Model Performance}
\label{subsubsec:mmsr_vote}
\begin{figure}[t]
  \centering
   \includegraphics[width=0.95\columnwidth]{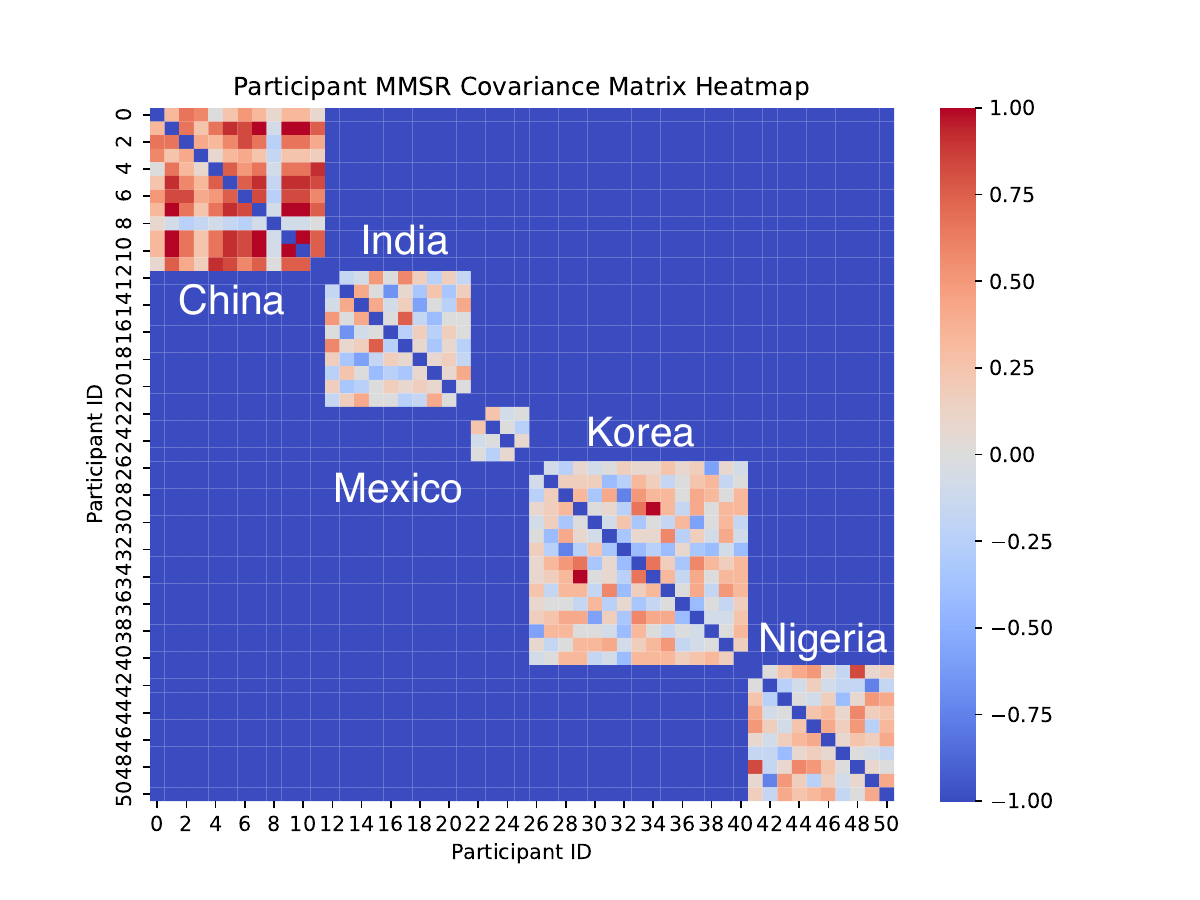}

   \caption{MMSR Covariance Matrix of Participant Response Agreement when evaluating combinations of Neural Network Models (e.g. contrastive), Survey Items (e.g. Offensiveness), and Country (e.g. Nigeria). Each row and column denotes a different person, where the more red the square, the more the participants agree with each other. -1 is maximum disagreement (or no mutual responses), and 1 means maximum agreement. Each of the big grouping rectangles represents data from a different country, ordered from top left to bottom right as China, India, Mexico, Korea, Nigeria.}
   \label{fig:mmsr_covariance_matrix}
   \vspace{-0.5cm}
\end{figure}

We quantitatively estimate the subjective perceived performance of each model using the crowd-kit
implementation of the Matrix Mean-Subsequence-Reduced (MMSR)
model, an established algorithm
for noisy label aggregation, followed by a weighted majority vote to aggregate labels across workers, and then a simple majority vote aggregating labels into rankings, thus MMSR+Vote. 
MMSR models the varying levels of participant expertise as a vector, which we frame as representing consensus alignment to account for the combination of knowledge and subjective perception we are measuring, and the equivalent term chosen by 
is ``skills''.
We abstract rankings in survey responses into unique binary pairwise comparison labels asking if the left image is perceived as better than the right image with respect to the given survey item.
An example of a binary comparison is when a respondent has indicated that it is true that an image given rank 1 is less offensive than the image given rank 2.
MMSR is provided with the participant, survey item, and abstracted response labels, then models the noisy label prediction problem as:
\begin{equation}
     \mathbb{E}\left[\frac{M}{M-1}\widetilde{C}-\frac{1}{M-1}\boldsymbol{1}\boldsymbol{1}^T\right]
     = \boldsymbol{s}\boldsymbol{s}^T,
\end{equation}
where $\widetilde{C}$ is the participant covariance matrix (Figure~\ref{fig:mmsr_covariance_matrix}), $M$ is the number of labels (true, false), and $boldsymbol{1}$ is a matrix filled with the width and height of $\widetilde{C}$.
We run 10k iterations of robust rank-one matrix completion with a stopping tolerance of 1e-10 to compute the initial label estimates.
We aggregate the binary labels with a weighted majority vote, and then estimate aggregate rankings from labels with a simple majority vote.

One limitation of MMSR evaluation is that it prioritizes a single consensus response.
This means a comparatively small quantity of insightful but marginalized perspectives might be undervalued in a manner undifferentiated from a small quantity of random responses.

\begin{figure*}
    \centering
    \includegraphics[width=\textwidth]{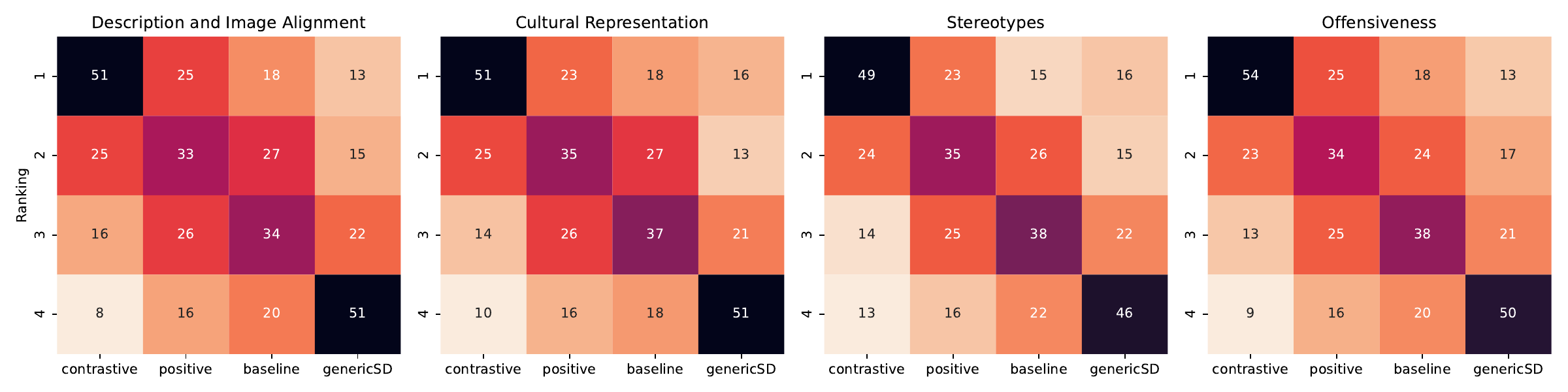}
    \caption{Counts of Participant-selected Model ranking for each survey item across all responses, more participants choosing a better rank value (lower number) is better. Our contrastive approach is selected for rank 1 most frequently across all survey evaluation items.}
    \label{fig:participant-rank-count}
\end{figure*}

\noindent We run MMSR under three configurations: 

\noindent \textbf{(1) Figure~\ref{fig:mmsr_overall} Overall Best Method:} where all rankings for all items and all countries are supplied together to estimate the best method across all responses.
This model found a participant consensus indicating our SCoFT+MPC method performs best, followed by SCoFT+MP, SCoFT+M, and finally Generic Stable Diffusion as the least preferred.

\noindent \textbf{(2) Figure~\ref{fig:mmsr_country} Overall best method by Country:} This model found a participant consensus in complete agreement with configuration 1, with the exception of India where the consensus agreement rated the SCoFT+MP method best, followed by SCoft+MPC, SCoFT+M, and Generic Stable Diffusion.

\noindent \textbf{(3) Figure~\ref{fig:mmsr_survey_item} Overall best method by Survey Item:} This model found overall participant ratings of models across each of the survey topics (Prompt to Image Alignment, Cultural Representation, Stereotyping, and Offensiveness) agree with configuration 1.

Figure~\ref{fig:mmsr_covariance_matrix} shows the MMSR Covariance matrix heatmap representing the strength of agreement between different participants.
Each row and column is a separate participant. The deep blue areas represent questions with zero overlap, as participants from each country were asked independent groups of questions. 
The lighter blue through dark red squares are the similarity of answers between ay two participants.
The squares from top left to bottom right represent China, India, Mexico, Korea, and Nigeria, respectively.
The covariance matrix for China is notably a darker red compared to all other countries, indicating a combination of a stronger agreement in combination with the greater number of questions answered per participant with experiences in China, on average.


Figure~\ref{fig:participant-rank-count} displays the counts of survey item responses (Table \ref{tab:survey_items}) for each ablation. Our contrastive approach SCoFT+MPC is consistently selected as the top-ranked choice across all survey evaluation items.

\textbf{Noisy Bradley Terry.} We quantitatively estimate the subjective perceived performance of each model using the crowd-kit
implementation of the Noisy Bradley Terry (NBT)
model, an algorithm
that performs noisy binary label aggregation for the purpose of further substantiating our existing results.
As Figure~\ref{fig:nbt_combined} shows, the NBT composite score of participant responses consistently ranks SCoFT+MPC method as the best, followed by SCoFT+MP, SCoFT+M, and finally Generic Stable Diffusion as the least preferred in every case, including the Overall rankings in Figure~\ref{fig:nbt_overall}, the rankings by Country in Figure~\ref{fig:nbt_country}, and the Rankings by Survey Item in Figure~\ref{fig:nbt_survey_item}.

Taken together, the reliable consistency of model rankings across multiple different evaluation methods, maximum participant count (Figure~\ref{fig:participant-rank-count}), and simple averaging, represent very strong quantitative evidence of the improvements constituted by our SCoFT+MPC method with respect to Description and Image Alignment, Cultural Representation, Stereotyping, and Offensiveness when compared to Stable Diffusion.

\section{Additional Qualitative Samples}
\label{sec:app_qualitative}
\subsection{Memorization Loss}
\begin{figure*}
    \centering
    \includegraphics[width=\textwidth]{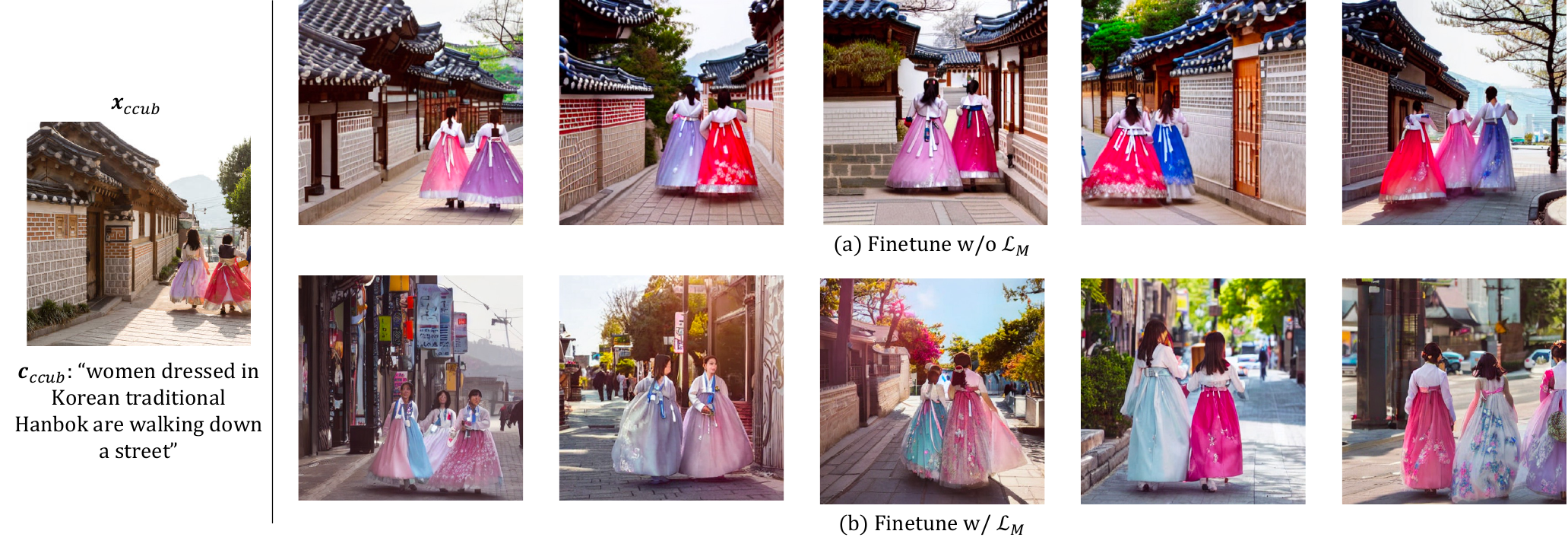}
    \caption{Additional qualitative example comparing training of 6000 steps with and without $\mathcal{L}_{M}$. The generated images are conditioned on the training text prompt: ``women dressed in Korean traditional Hanbok are walking down a street". }
    \label{fig:app_diverse1}
\end{figure*}

\begin{figure*}
    \centering
    \includegraphics[width=\textwidth]{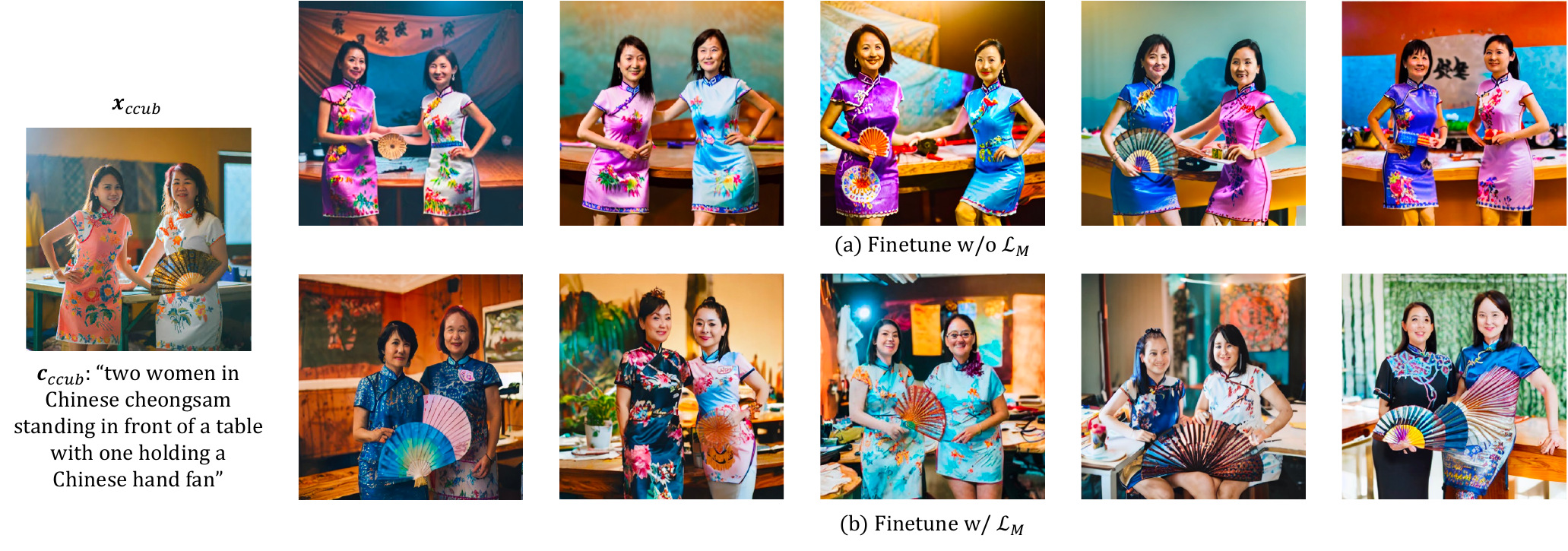}
    \caption{Additional qualitative example comparing training of 6000 steps with and without $\mathcal{L}_{M}$. The generated images are conditioned on the training text prompt: ``two women in Chinese cheongsam standing in front of a table with one holding a Chinese hand fan".}
    \label{fig:app_diverse3}
\end{figure*}

\begin{figure*}
    \centering
    \includegraphics[width=\textwidth]{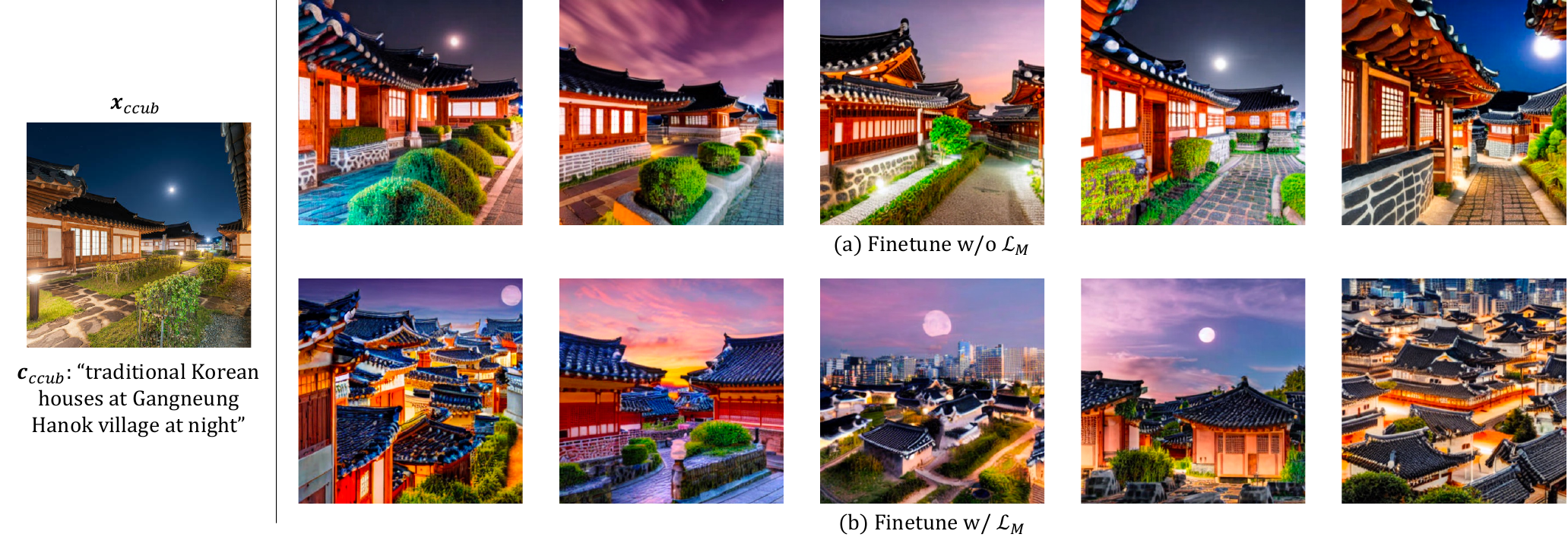}
    \caption{Additional qualitative example comparing training of 6000 steps with and without $\mathcal{L}_{M}$. The generated images are conditioned on the training text prompt: ``traditional Korean houses at Gangneung Hanok village at night".}
    \label{fig:app_diverse2}
\end{figure*}

We present more examples comparing fine-tuning Stable Diffusion with CCUB using $\mathcal{L}_{LDM}$ with and without memorization loss ($\mathcal{L}_{M}$) in Figures \ref{fig:app_diverse1}, \ref{fig:app_diverse3}, and \ref{fig:app_diverse2}. 
The addition of $\mathcal{L}_M$ prevents the model from generating images similar to the training images given a training text prompt. This is an important property when fine-tuning on a small dataset, e.g., CCUB.


\subsection{Ablation on Self-Contrastive Perceptual Loss}
\begin{figure*}
    \centering
    \includegraphics[width=0.95\linewidth]{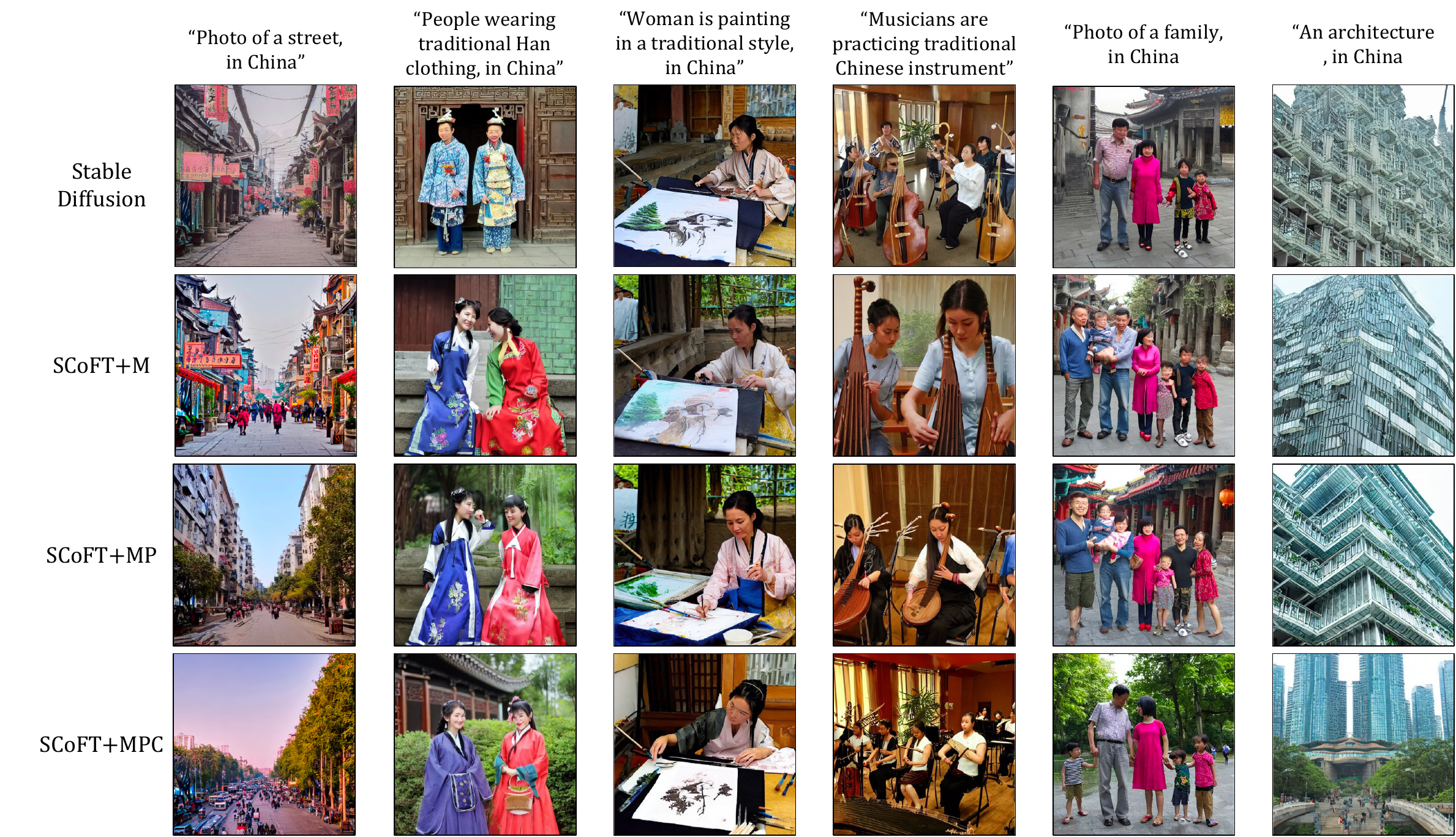}
    \caption{\textbf{Ablation on SCoFT for Chinese culture.} We present additional qualitative examples for fine-tuning on the CCUB Chinese dataset using different losses. Generic Stable Diffusion often results in stereotypes and misrepresentations of Chinese culture. In contrast, our SCoFT+MPC approach achieves superior results, generating accurate and less offensive images.}
    \label{fig:app_ablation_china}
\end{figure*}

\begin{figure*}
    \centering
    \includegraphics[width=0.95\linewidth]{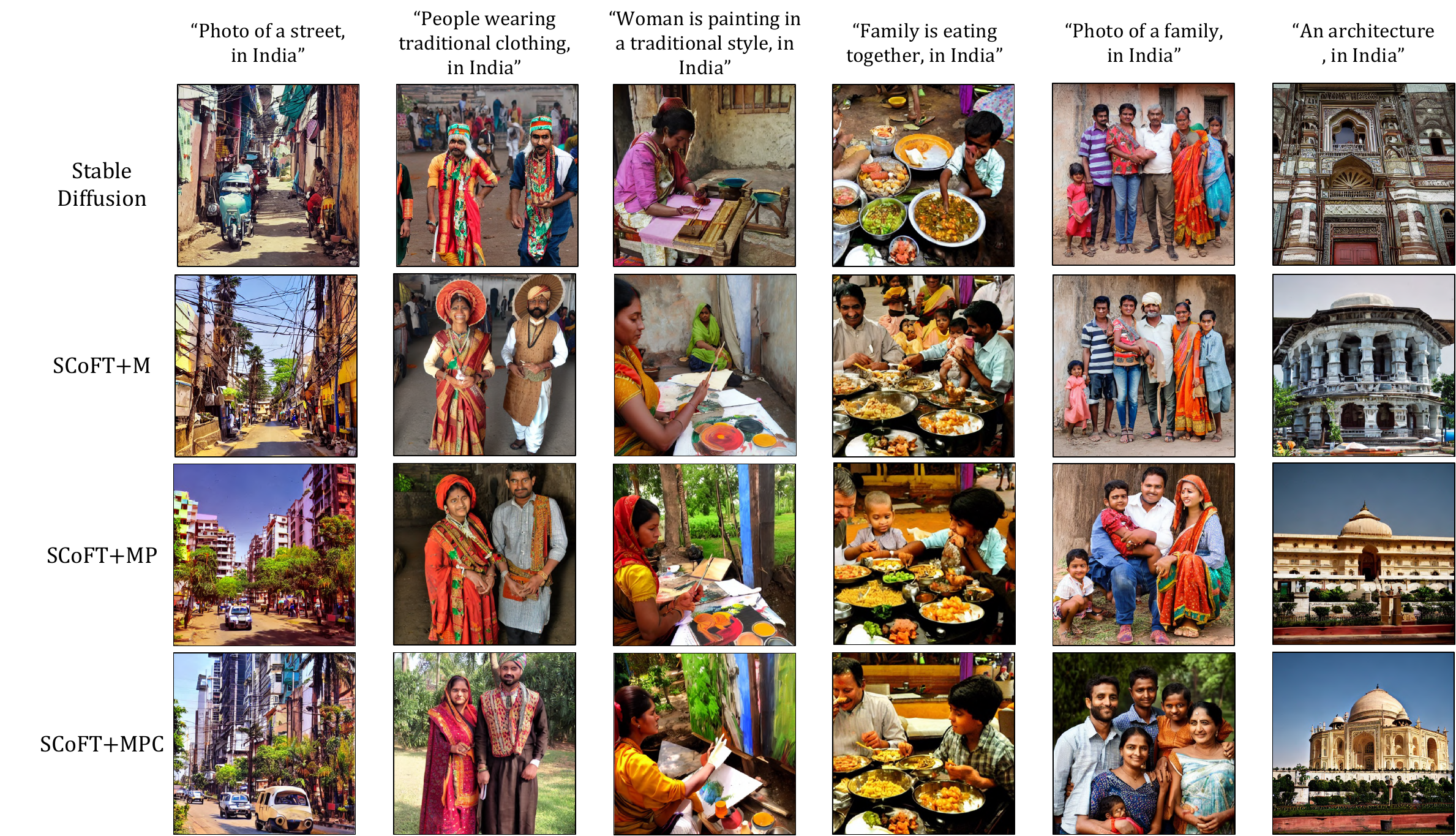}
    \caption{\textbf{Ablation on SCoFT for Indian culture.} We present additional qualitative examples for fine-tuning on the CCUB India dataset using different losses. Generic Stable Diffusion often results in stereotypes and misrepresentations of Indian culture. In contrast, our SCoFT+MPC approach achieves superior results, generating accurate and less offensive images.}
    \label{fig:app_ablation_india}
\end{figure*}

In Figures~\ref{fig:app_ablation_china} and~\ref{fig:app_ablation_india}, we present additional ablations on SCoFT, showcasing the impact of different losses on the CCUB datasets for Chinese and Indian cultures.  SCoFT improves generated images compared to conventional fine-tuning by reducing stereotypes such as old, poor looking regions and adhering to culturally important aspects such as art tools, clothing, and architectual styles.

\begin{figure*}
    \centering
    \includegraphics[width=0.95\linewidth]{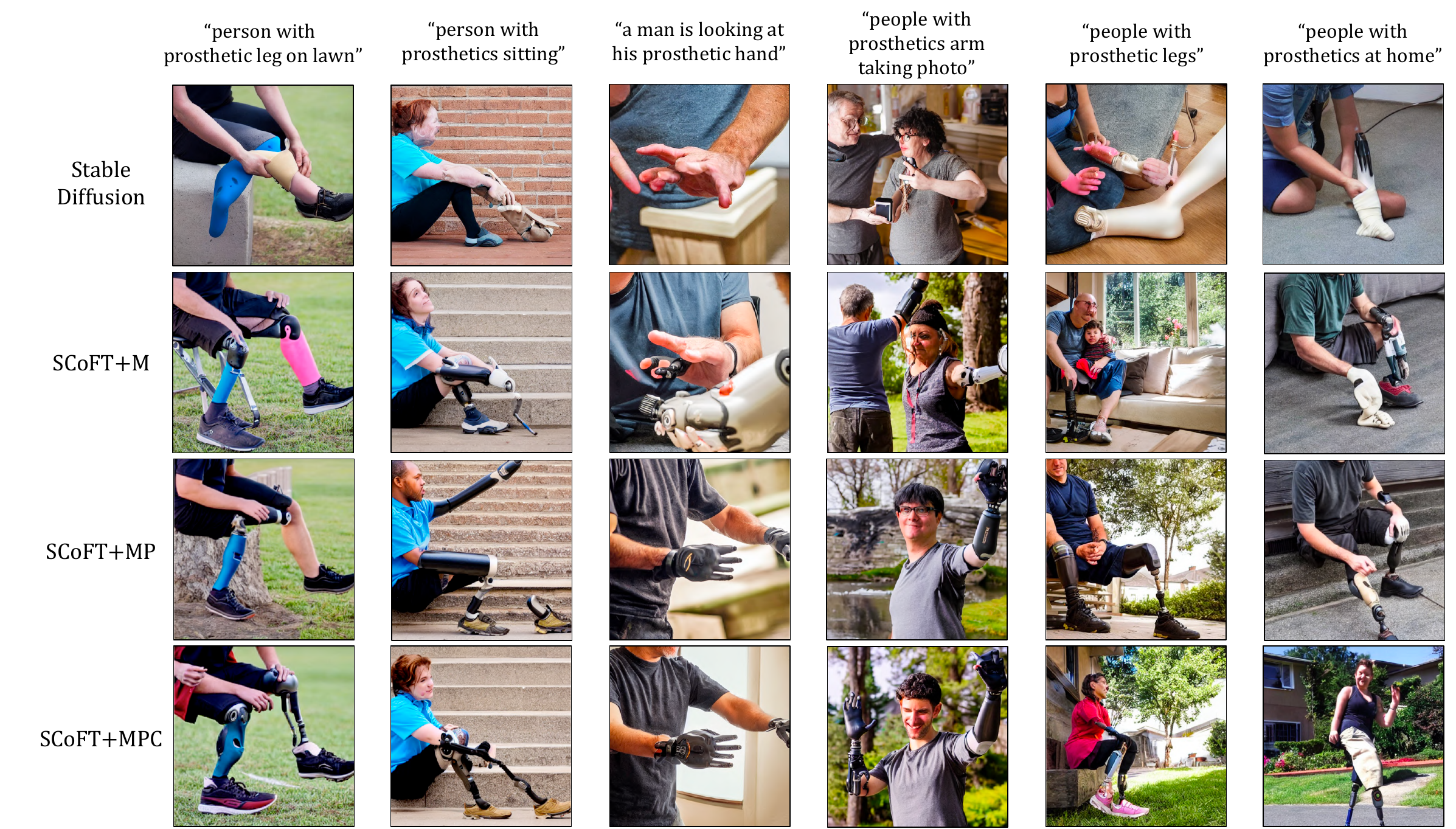}
    \caption{\textbf{Ablation on SCoFT for prosthetic dataset.} We provide additional qualitative examples for fine-tuning on the prosthetic dataset using different losses. Generic Stable Diffusion struggles to generate accurate representations for people with prosthetics. In contrast, our SCoFT+MPC approach achieves superior results, offering accurate representations.}
    \label{fig:app_prosthetics}
\end{figure*}

In Figure~\ref{fig:app_prosthetics}, we validate the efficacy of each loss in our SCoFT framework using an internal prosthetics dataset. Our findings indicate that generic Stable Diffusion tends to introduce inherent biases in representations of the prosthetic using community and often generates inaccurate images. In contrast, our SCoFT+MPC approach consistently produces accurate representations. Moreover, we demonstrate the versatility of SCoFT by applying it to other fine-tuning domains, such as the prosthetics dataset, where it proves effective in generating more accurate images. We hope to perform future work in demonstrating SCoFT's abilities to improve generated images for many communities harmed by bias and inaccurate representation.

\subsection{More Qualitative Examples}

\begin{figure*}
    \centering
    \includegraphics[width=0.9\linewidth]{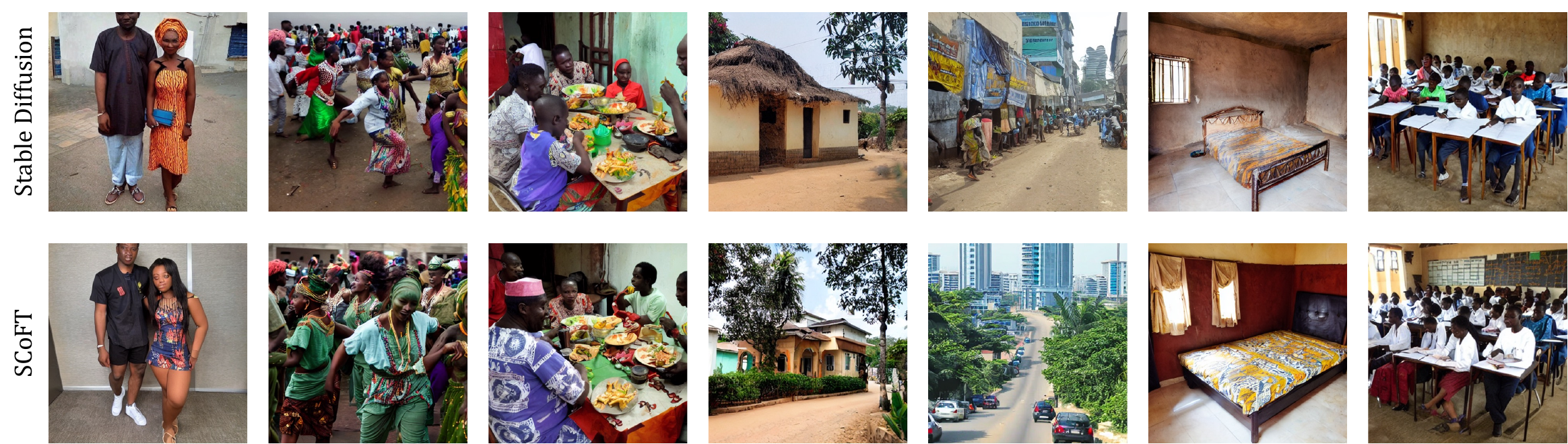}
    \caption{\textbf{Additional qualitative examples comparing Stable Diffusion with our SCoFT model for Nigerian culture.} The text prompts are ``Nigerian people in casual clothing nowadays", ``dancers are performing for a crowd, in Nigeria", ``family is eating together, in Nigeria", ``photo of a house, in Nigeria", ``photo of a street, in Nigeria", ``photo of a bedroom, in Nigeria", ``student studying in the classroom, in Nigeria".}
    \label{fig:app_show_nigeria}
\end{figure*}

\begin{figure*}
    \centering
    \includegraphics[width=0.9\linewidth]{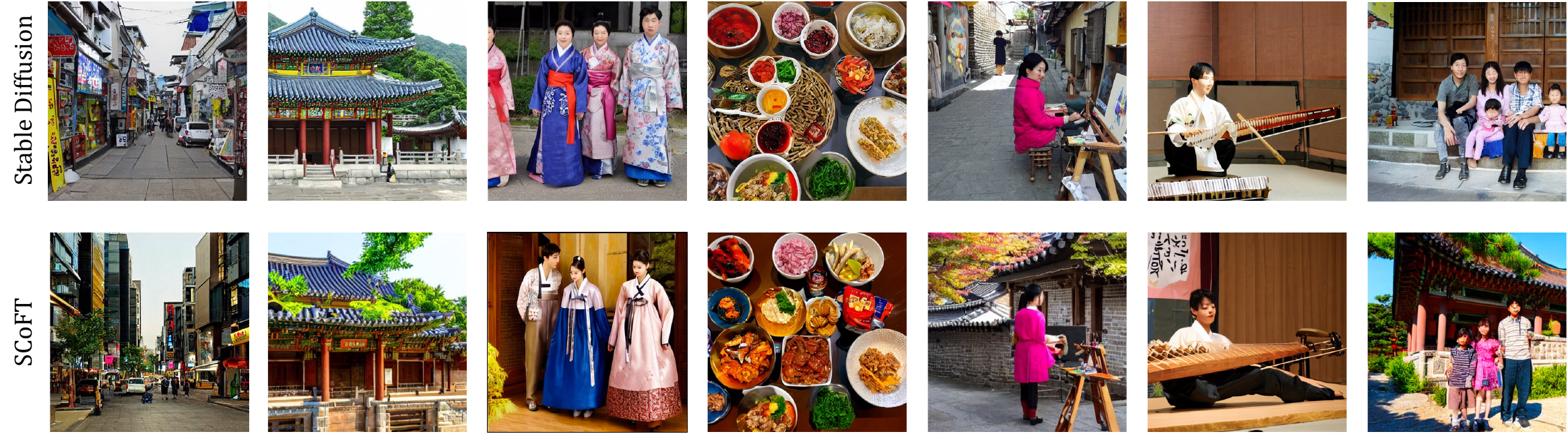}
    \caption{\textbf{Additional qualitative examples comparing Stable Diffusion with our SCoFT model for Korean culture.} The text prompts are ``photo of a street, in Korea", ``photo of a traditional building, in Korea", ``people wearing traditional clothing, in Korea", ``a table of food in Korea", ``a woman is painting in a traditional style, in Korea", ``musician performing Korean traditional instrument", ``photo of a family, in Korea".}
    \label{fig:app_show_korea}
\end{figure*}

\begin{figure*}
    \centering
    \includegraphics[width=0.9\linewidth]{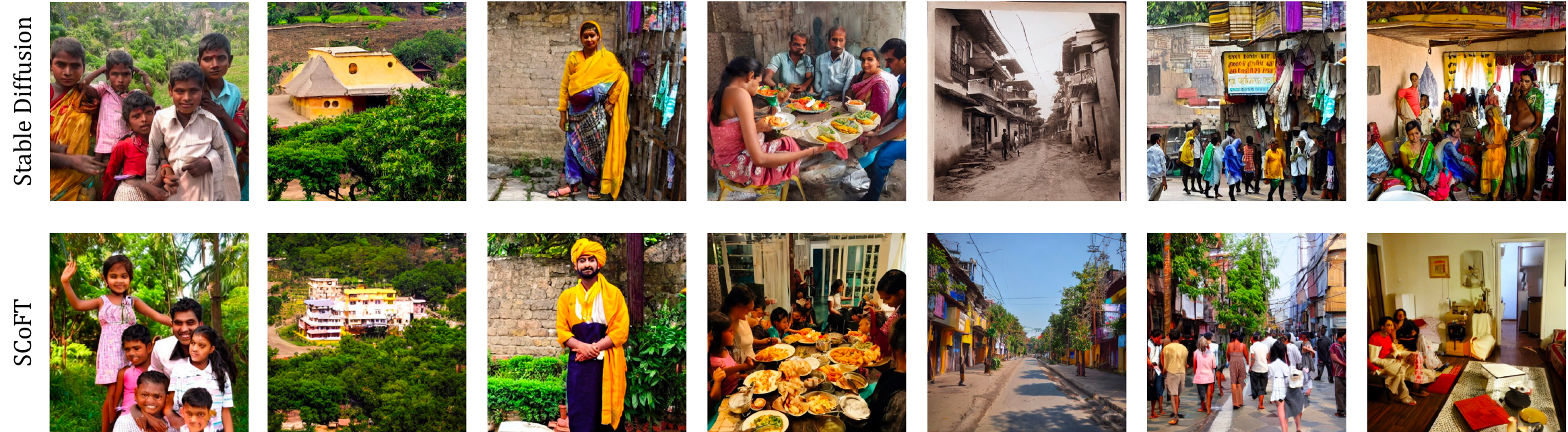}
    \caption{\textbf{Additional qualitative examples comparing Stable Diffusion with our SCoFT model for Indian culture.} The text prompts are ``photo of children in India", ``photo of a house, in India", ``people wearing traditional clothing, in India", ``family is eating together, in India", ``photo of a street, in India", ``people walking on the street, in India", ``people inside their house, in India".}
    \label{fig:app_show_india}
\end{figure*}

\begin{figure*}
    \centering
    \includegraphics[width=0.9\linewidth]{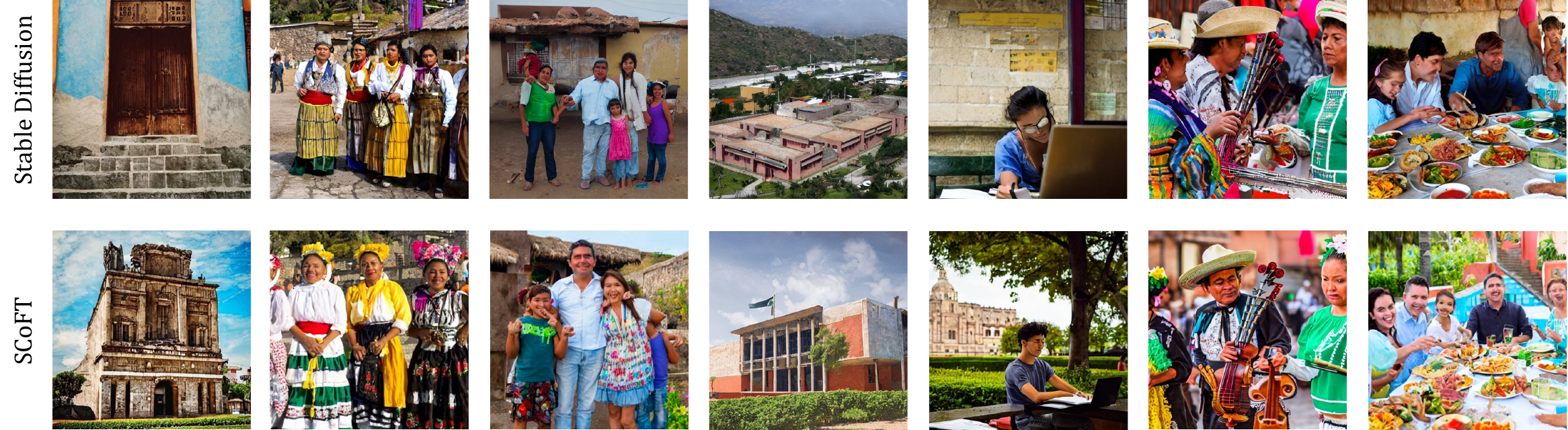}
    \caption{\textbf{Additional qualitative examples comparing Stable Diffusion with our SCoFT model for Mexican culture.} The text prompts are ``photo of a building, in Mexico", ``people wearing traditional clothing, in Mexico", ``photo of a family, in Mexico", ``photo of a school, in Mexico", ``university student studying, in Mexico", ``people performing traditional music instrument, in Mexico", ``family is eating together, in Mexico".}
    \label{fig:app_show_mexico}
\end{figure*}

\begin{figure*}
    \centering
    \includegraphics[width=0.9\linewidth]{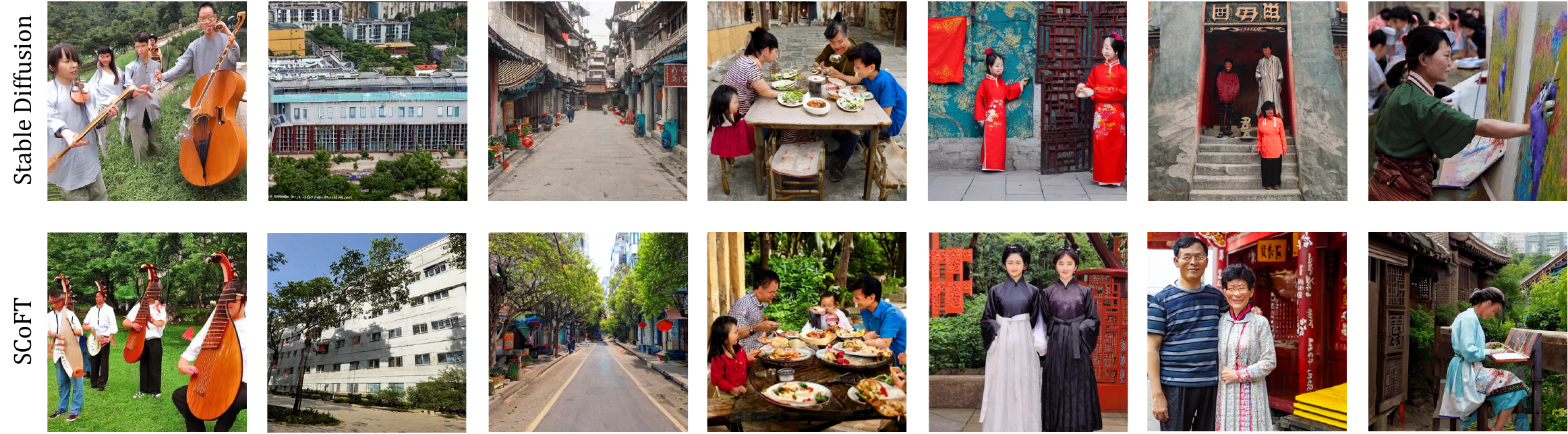}
    \caption{\textbf{Additional qualitative examples comparing Stable Diffusion with our SCoFT model for Chinese culture.} The text prompts are ``people are performing traditional instrument, in China", ``photo of a school, in China", ``photo of a street, in China", ``family is eating together, in China", ``two girls wearing Chinese traditional Han dress", ``a man and a woman, in China", ``woman is painting in a traditional style, in China".}
    \label{fig:app_show_china}
\end{figure*}

In Figures~\ref{fig:app_show_nigeria} to~\ref{fig:app_show_china}, we present additional qualitative examples illustrating how our SCoFT model outperforms the original Stable Diffusion in cultural understanding and the ability to generate less offensive images for various cultural concepts. Results are showcased across Nigeria, Korea, India, Mexico, and China. Furthermore, our models exhibit good performance for text prompts beyond our nine cultural categories, such as ``photo of a bedroom", ``students are studying in the classroom".

\section{Additional Limitations}
\label{sec:app_limitation}
This work has the potential to shift the way that image generators operate at achievable costs to ensure that several categories of harm from `AI' generated models are mitigated, while the generated images become much more realistic and representative of the AI-generated images that populations want around the world. Our proposed methods have potential applications in other domains such as reducing the risk of copyright infringement, better respecting cultural and community-defined boundaries, and addressing offensiveness across a broader range of identity characteristics and other criteria. Additionally, this work carries several risks, for example, the algorithm can easily be inverted to generate the most problematic images possible. While our approach works toward collecting and training datasets in a respectful manner, we do not address the nonconsensual use of images for training the baseline Stable Diffusion models we use as a starting point.

\clearpage

